\documentclass[journal, onecolumn]{IEEEtran}

\usepackage{amsmath}
\usepackage{amssymb}

\usepackage{lineno}

\usepackage{bm}
\usepackage{subcaption}
\usepackage{algorithm}
\usepackage{algpseudocode}
\usepackage{multirow}
\usepackage{hhline}

\usepackage{color}
\usepackage{xcolor,colortbl}
\usepackage[normalem]{ulem}

\usepackage{graphicx}

\definecolor{Gray}{gray}{0.85}

\begin{document}

\title{Generation of BIM data based on the automatic detection, identification and localization of lamps in buildings}
\author{Francisco Troncoso-Pastoriza, Pablo Egu\'{i}a-Oller, Rebeca P. Díaz-Redondo, Enrique Granada-\'{A}lvarez
\thanks{Francisco Troncoso-Pastoriza, Pablo Egu\'{i}a-Oller and Enrique Granada-\'{A}lvarez are with School of Industrial Engineering, University of Vigo, Campus Universitario, 36310 Vigo (Spain)}
\thanks{Rebeca P. Díaz-Redondo (rebeca@det.uvigo.es) is with School of Telecommunication Engineering, University of Vigo, Campus Universitario, 36310 Vigo (Spain)}
}

\maketitle

\begin{abstract}
In this paper we introduce a method that supports the detection, identification and localization of lamps in a building, with the main goal of automatically feeding its energy model by means of Building Information Modeling (BIM) methods. The proposed method, thus, provides useful information to apply energy-saving strategies to reduce energy consumption in the building sector through the correct management of the lighting infrastructure. Based on the unique geometry and brightness of lamps and the use of only greyscale images, our methodology is able to obtain accurate results despite its low computational needs, resulting in near-real-time processing. The main novelty is that the focus of the candidate search is not over the entire image but instead only on a limited region that summarizes the specific characteristics of the lamp. The information obtained from our approach was used on the Green Building XML Schema to illustrate the automatic generation of BIM data from the results of the algorithm.
\end{abstract}


\section{Introduction}
\label{intro}

The reduction of energy consumption is a key factor for increasing the competitiveness not only in the industrial sector but also in public or private administration buildings. Advancement towards a more efficient future model with better environmental performance is the goal, and with this aim, different initiatives, private and public, have arisen, notably due to the efforts of the European Union under the H2020 initiative~\cite{H2020}. In fact, this programme specifies the following objectives: (i) a reduction in the consumption of primary energy in the European Union by 20\%, (ii) reduction in the greenhouse gas emissions by another 20\%, and (iii) an increase in the contribution of renewable energies to 20\% of the total consumption.

Contrary to appearances, the energy consumption of the building sector (private, mercantile and service) represents approximately 40\% of the total energy consumption worldwide and is a more than significant source of greenhouse gas emissions. Moreover, its growth trend is not expected to decrease in the short term~\cite{Lombard}. However, more adequate management of lighting could be key to the reduction of energy consumption, especially in office buildings~\cite{Soori}. Thus, making the right decisions regarding the location, type and state of lamps is essential to appropriate energy-saving strategies.

Consequently, knowing the real lighting conditions of a building is the first step of any initiative to reduce its energy consumption. Necessary information concerning these conditions includes (i) a detailed lighting inventory and (ii) the state of the lamps (on/off). There have been different approaches to automatically perform both processes, like the work of Vilari\~{n}o \emph{et al.}~\cite{Vilariño}. In this work the results are very accurate, but it has two main drawbacks: on the one hand, automation requires costly infrastructure; on the other hand, it also requires high computational costs and remarkable processing time, which prevents real-time operation.

Building Information Modeling (BIM) is a collaborative work methodology for the creation and management of a construction project. The objective is to centralize all the project information in a digital model, which conforms a big database that allows the management of all the elements of the infrastructure throughout its entire lifecycle.
It is an evolution from the traditional design systems based on planes, incorporating geometric information, times, costs, environment and maintenance. One of the main goals of the BIM methodology is to work efficiently, trying to optimize all the activities that make up a project and then reduce the duration and increase the productivity; this philosophy was also extended to energy efficiency. A recent study of this methodology can be found in~\cite{Zanni}. We use BIM as the foundation for a seamless integration of the information obtained with the algorithms presented in this work in the construction project.

Thanks to the increasing availability of RGB-D sensors or stereoscopic cameras and laser triangulation systems, many computer vision systems currently focus on depth data. However, this option usually imposes a more restrictive viewing range for correct operation. Moreover, the associated costs are still higher than those using simple cameras. Hence, edge-based image algorithms still perform better for object detection and location in the majority of cases~\cite{D2CO}. Actually, edge-based systems often perform an exhaustive search over the entire image, conducting template matching in the large space of possible rigid transformations. Although there are efficient methods to reduce the computation time, the number of candidates that must be tested to detect and locate an object is still large, involving complex and costly computations over the entire image~\cite{D2CO, Liu}. Additionally, some of the approaches are not even robust to depth variations outside of a limited range~\cite{Liu}.

The rest of the paper is structured as follows: Section~\ref{sec:sota} describes some related work in the field of object detection, location and pose (the combination of position and orientation) estimation; Section~\ref{sec:method} details the proposed method for lamp detection; the results of our approach are presented in Section~\ref{sec:results} to evaluate its performance; and finally, Section~\ref{sec:conclusions} provides some conclusions drawn from the results obtained.

\section{Related Work}
\label{sec:sota}

The use of BIM in building management and building design is the focus of much research nowadays, including building lighting information.
Jalaei and Jrade~\cite{Jalaei} state that the perimeter lighting of a building must be optimized and describe a methodology that integrates BIM with the LEED certification system.
Rahmani \emph{et al.}~\cite{Rahmani} concludes that building professionals are expected to improve energy performance of their design, integrating BIM with simulation engines and exposing electric and daylighting results directly on the BIM model.
Tronchin \emph{et al.}~\cite{Tronchin} remark that building data management systems need to be connected to BIM technology, including lighting as an important internal gain.
Soust-Verdaguer \emph{et al.}~\cite{Soust} use BIM and Life Cycle Assessment to quantify environmental impacts in the building sector highlighting the interoperability among software applications, like daylighting simulation.
Gerrish \emph{et al.}~\cite{Gerrish} evaluated the potential of using BIM as a tool to support visualization and management of building performance, where the space characteristics (like expected lighting) give the performance prediction.
Habibi~\cite{Habibi} combine simulation methods and optimization tools with BIM models to improve the performance of the building, pointing out that it is essential to collect all the possible data about the lighting equipment.

Methods for object detection and pose estimation are usually classified into the following two main categories: (i) image-based and (ii) model-based techniques. The former, image-based, extracts distinctive local features from different perspectives to perform reliable matching between different views of the object. Viksten \emph{et al.}~\cite{Viksten} perform a comparison among 6 degree-of-freedom pose estimation system for fourteen types of local descriptors. Although this approach obtains good results for textured objects, its performance is not as good when applied to textureless objects and non-Lambertian materials, such as glass or metal. The latter, model-based techniques use a 3D CAD model of the target object, extracting shape information for different viewpoints. This is a well-suited technique for textureless objects, being clearly superior to the traditional texture-based description for this type of target~\cite{Tombari}.

The Chamfer Distance Transform, together with its variations, has been successfully used in model-based methods for edge-based matching. This technique was first proposed by Barrow \emph{et al.}~\cite{Barrow} and seconded and improved by Borgefors~\cite{Borgefors}. Later, Choi and Christensen~\cite{Choi} used chamfer matching inside a particle-filtering framework for textureless object detection and tracking. Shotton \emph{et al.}~\cite{Shotton} proposed a novel formulation of chamfer matching by introducing an additional channel for edge point orientations called Oriented Chamfer Matching (OCM). This scheme was also used by Cai \emph{et al.}~\cite{Cai} for hypothesis verification. Danielsson \emph{et al.}~\cite{Danielsson} proposed the use of multi-local features to perform object category detection via a voting scheme based on computing the distance transform maps for a set of discretized orientations. Similar to the latter, Damen \emph{et al.}~\cite{Damen0, Damen1} proposed a simple method based on constellations of edgelets, which was later improved in~\cite{Damen2}. Liu \emph{et al.} extended the idea of~\cite{Danielsson} with the Fast Directional Chamfer Matching (FDCM)~\cite{Liu}, which performs a 3D distance transform over a joint location/orientation space. Recently, Imperioli and Pretto introduced an edge-based registration algorithm based on the work of Liu \emph{et al.} called Direct Directional Chamfer Optimization (D\textsuperscript{2}CO)~\cite{D2CO}, which refines the object position employing a non-linear optimization procedure, where the cost being minimized is extracted directly from the 3D image tensor.

\section{Methodology}
\label{sec:method}

Our proposal addresses the problem of knowing the real lighting conditions of a building (inventory and state), focusing on the detection, location and identification of lamps with the aim of feeding its energy model. Our methodology is based on the following two relevant aspects: (i) the unique geometric and brightness of lamps and (ii) the use of greyscale images. This allows for a reduction of the matching process to identify a lamp to only a few candidates per image, which entails a great reduction of the computation costs. Consequently, we propose a fast solution (near-real-time processing) with accurate results.

The methodology consists of the following three stages for each image. First, some pose candidates are extracted from the input greyscale image in a fast and efficient manner, discarding false positives as soon as possible. After that, several poses are evaluated for each lamp model based on the initial candidates, and only the best are selected, identifying the correct lamp models. Finally, each surviving pose is refined, and a score is given. Our method leverages the Fast Directional Chamfer Distance~\cite{Liu} for candidate and model selection and the Direct Directional Chamfer Optimization (D\textsuperscript{2}CO)~\cite{D2CO} for pose refinement and scoring, while a novel method for fast candidate search is introduced.

Apart from the three aforementioned stages of our approach, the methodology also needs an additional stage, {\em Model registration} (Section~\ref{sect:modelRegistration}), whose aim is to obtain the model templates that will eventually be compared to locate and identify the lamps.

\begin{figure*}[h]
\centering
\includegraphics[width=0.7\linewidth]{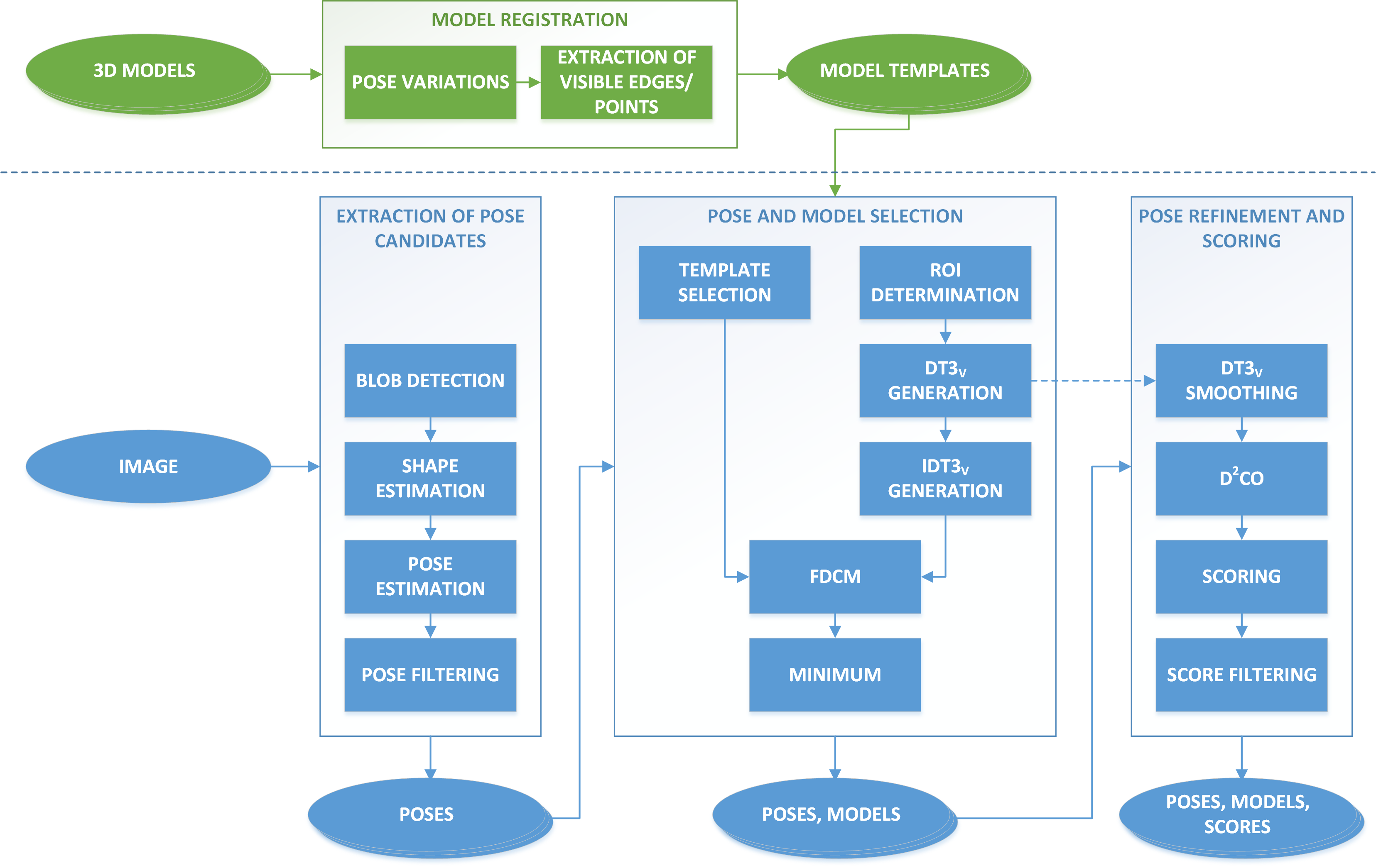}
\caption{Outline of the detection and location method.}
\label{fig:method}
\end{figure*}

Figure~\ref{fig:method} shows a global overview of the approach and the three main phases of the image processing pipeline: {\em Extraction of pose candidates} (Section~\ref{sect:extractionPoseCandidates}), {\em Pose and model selection} (Section~\ref{sect:poseModelSelection}) and {\em Pose refinement and scoring} (Section~\ref{sect:poseRefinementScoring}).

Moreover, to apply this method not only to isolated images but also to video frames, we perform some calculations explained in Section~\ref{sect:3DPoseProcessing}. Finally, in Section~\ref{sect:automaticInclusionLighting}, we show how to include the gathered information in the thermal model of the building.

We have applied the method to a set of 51 images extracted from three sequences recorded in different environments. The selection of images was done based on the camera pose to avoid including samples with very similar conditions: from the sequences we have never included images whose camera pose differ less than 50 cm in position and 20$^\circ$ in orientation with respect to any other image in the dataset.

\subsection{Model registration}
\label{sect:modelRegistration}

The model registration consist on the process of generating a set of templates from a 3D CAD model of each lamp to later match against an image. This procedure could be done directly while processing the image just with the 3D CAD model, but generating the template database beforehand helps reduce the processing time of the main method.

Therefore, we discretize the possible camera views in the 6D pose space. We can reduce the possible space based on the lamp type: for example, a ceiling lamp only needs views from the light surface (``southern'') hemisphere.

For each one of these camera poses, we have to generate a template with the edge information that would be visible in an image taken from such point of view. Hence, we need to extract the visible edges of the model and project them to the 2D camera space.

The visible edge extraction is a two-step process. In the first step we need to select the edges that are either sharp or part of the outline from the current perspective. This means that an edge between two faces with normals $\vec{n}_a$ and $\vec{n}_b$ is visible if it satisfies one of the following conditions:
\begin{itemize}
\item It is sufficiently sharp to be noticeable:
\begin{align}
\langle \vec{n}_a, \vec{n}_b \rangle < V,
\end{align}
with $V$ a visibility threshold ($V = 0.92$ in our experiments),
\item It is part of the outline of the object for the current viewpoint:
\begin{align}
\langle \vec{n}_a, \vec{c}_z \rangle \cdot \langle \vec{n}_b, \vec{c}_z \rangle \leq 0,
\end{align}
with $\vec{c}_z$ the viewing direction of the camera.
\end{itemize}

Even if an edge meets this first requirement, it can still be completely or partially occluded. The second step consist in filtering out the occluding parts. For this purpose, we discretize the edges, extracting points with a step of 1-2 mm. For each one of these points, we test if it is occluded based on the OpenGL~\cite{OpenGL} depth buffer. To this end, the OpenGL projection matrix is set to the intrinsic camera parameters of the camera used in the experiments, while the modelview matrix is set to the object pose.

Once we have the final visible points after the two tests, we can recompose the edges and project them to the 2D camera space. We store edges for the FDCM and the points with orientations for the D\textsuperscript{2}CO.

\subsection{Extraction of pose candidates}
\label{sect:extractionPoseCandidates}

The first stage, the process of extracting pose candidates, involves (i) the detection of blobs of bright areas in the image, followed by (ii) an estimation of the shape of the light surface, (iii) a pose estimation based on the hypothesized shape and (iv) some filters to reject bright areas not corresponding to a light source. 

\subsubsection{Blob detection}

\begin{figure}[h]
\centering
\includegraphics[width=0.8\linewidth]{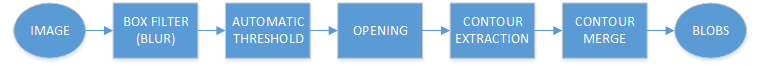}
\caption{Blob detection process.}
\label{fig:blobs}
\end{figure}

We use the method depicted in Figure~\ref{fig:blobs} to obtain blobs of possible light sources over a greyscale image (the input). First, the image is smoothed, by means of a normalized box filter with a kernel of size $3\times3$. Subsequently, a thresholding operation is applied to the smoothed image, resulting in a binary image. Later, an opening is performed by eroding and dilating the binary image. Then, the contours are extracted using the method presented by Suzuki and Abe~\cite{Suzuki}. Finally, adjacent contours are merged, computing the convex hull for each group of close blobs.

The lighting conditions for each image strongly depend on the context. Consequently, the threshold value for detecting bright areas must be dynamically selected and totally conditioned by the image being processed. Because the brightness properties of the image can be analysed via its histogram, we apply a normalization to the histogram values $f[k]$ of an image of size $m \times n$ as follows:
\begin{align}
\mathcal{F}[k] &= f[k] \cdot 256 \mathbin{/} (m \cdot n) .
\end{align}

\begin{figure*}[h]
\centering
\begin{subfigure}{0.151\textwidth}
\centering
\begin{subfigure}{\textwidth}
\centering
\includegraphics[width=\textwidth]{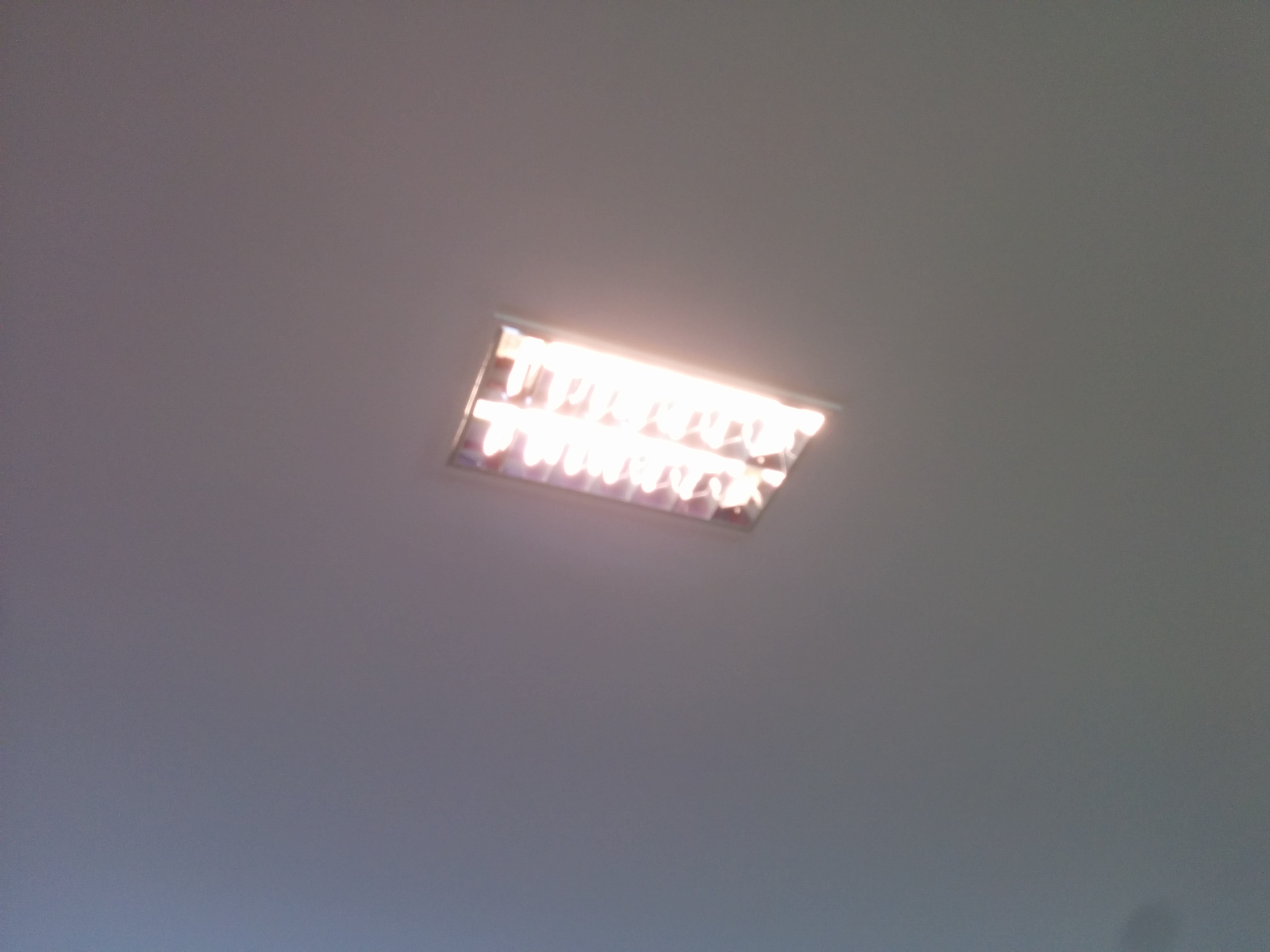}
\vspace*{0.25mm}
\end{subfigure}
\begin{subfigure}{\textwidth}
\centering
\includegraphics[width=\textwidth]{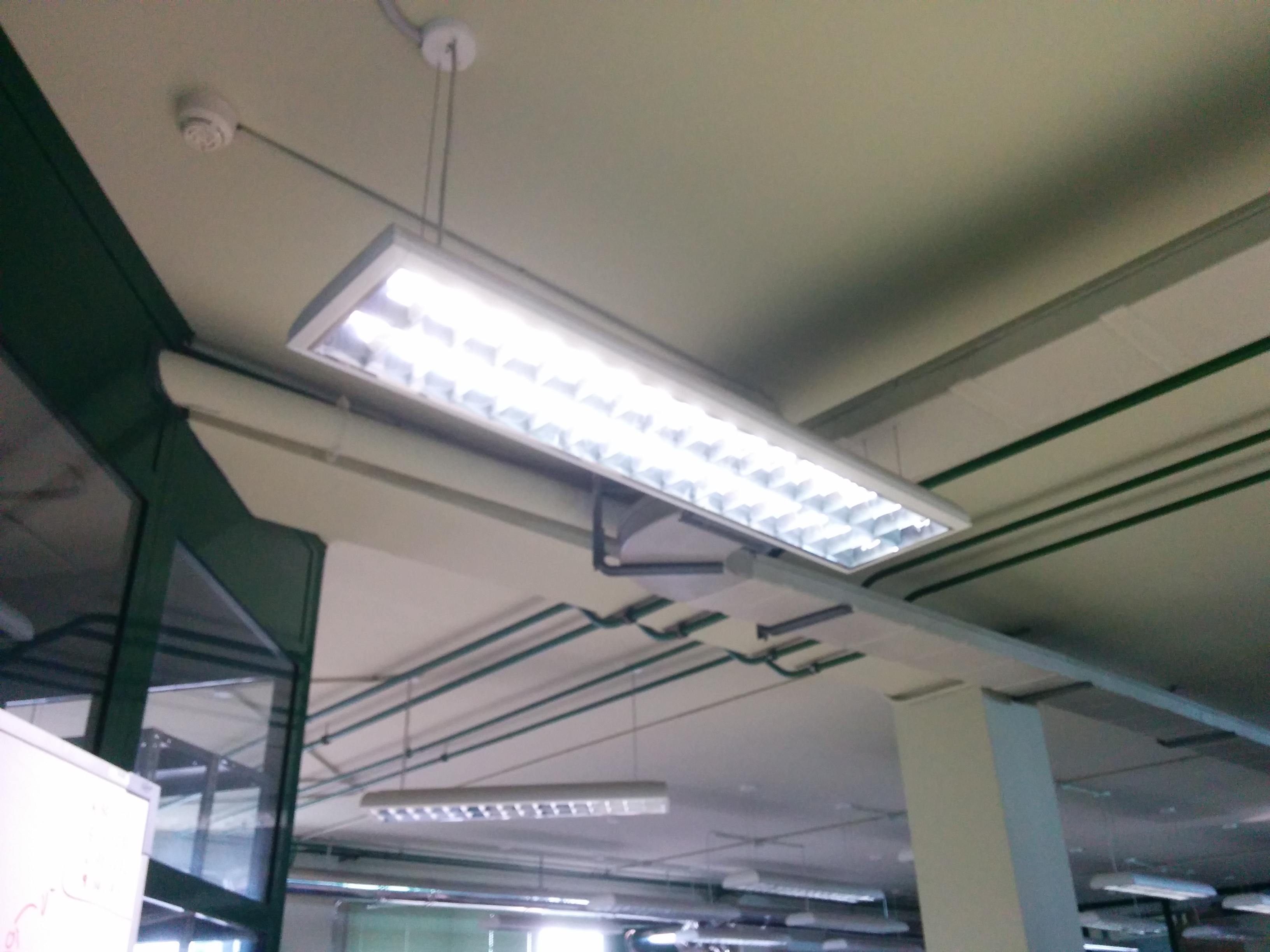}
\vspace*{0.25mm}
\end{subfigure}
\begin{subfigure}{\textwidth}
\centering
\includegraphics[width=\textwidth]{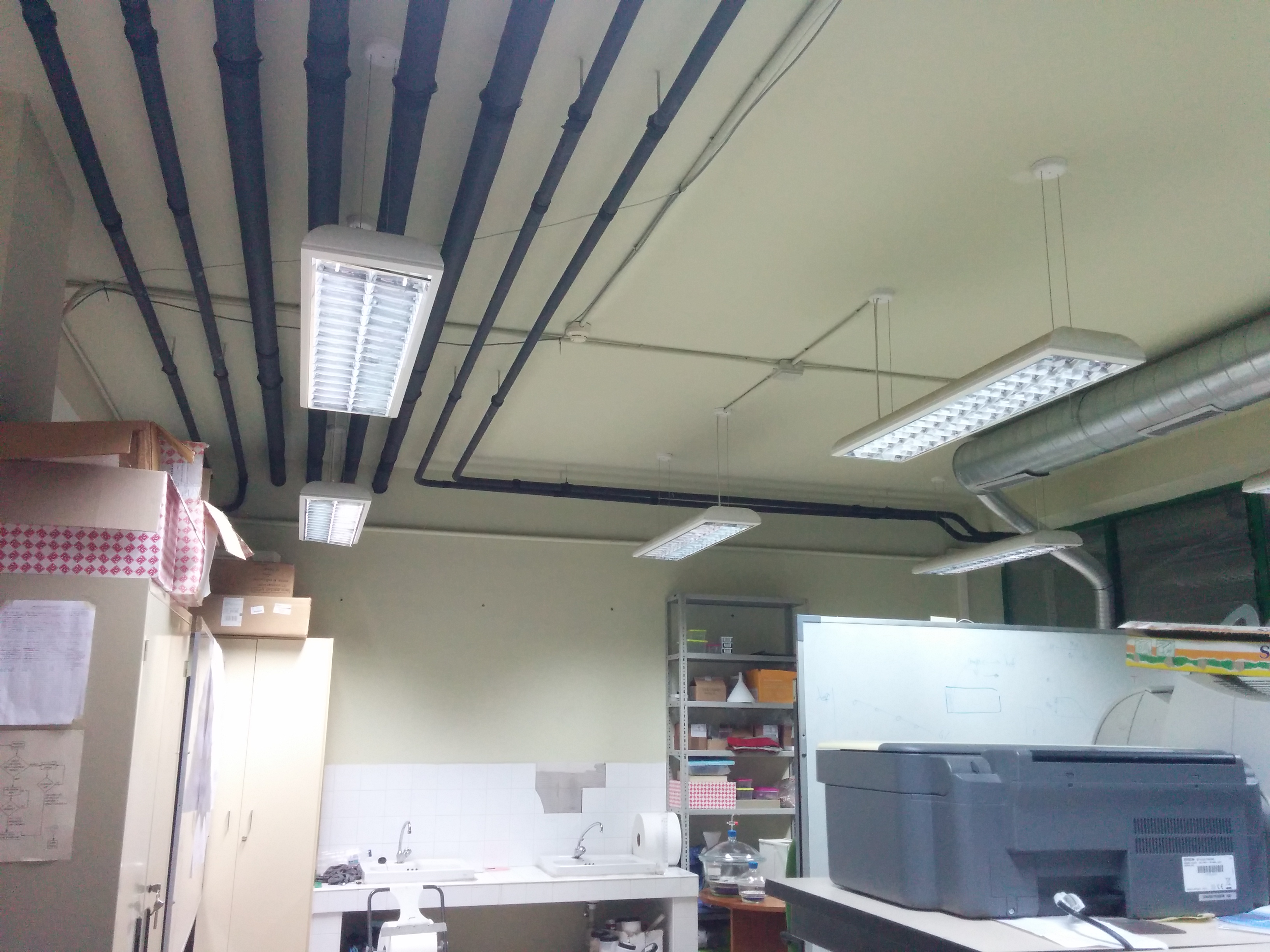}
\vspace*{0.25mm}
\end{subfigure}
\begin{subfigure}{\textwidth}
\centering
\includegraphics[width=\textwidth]{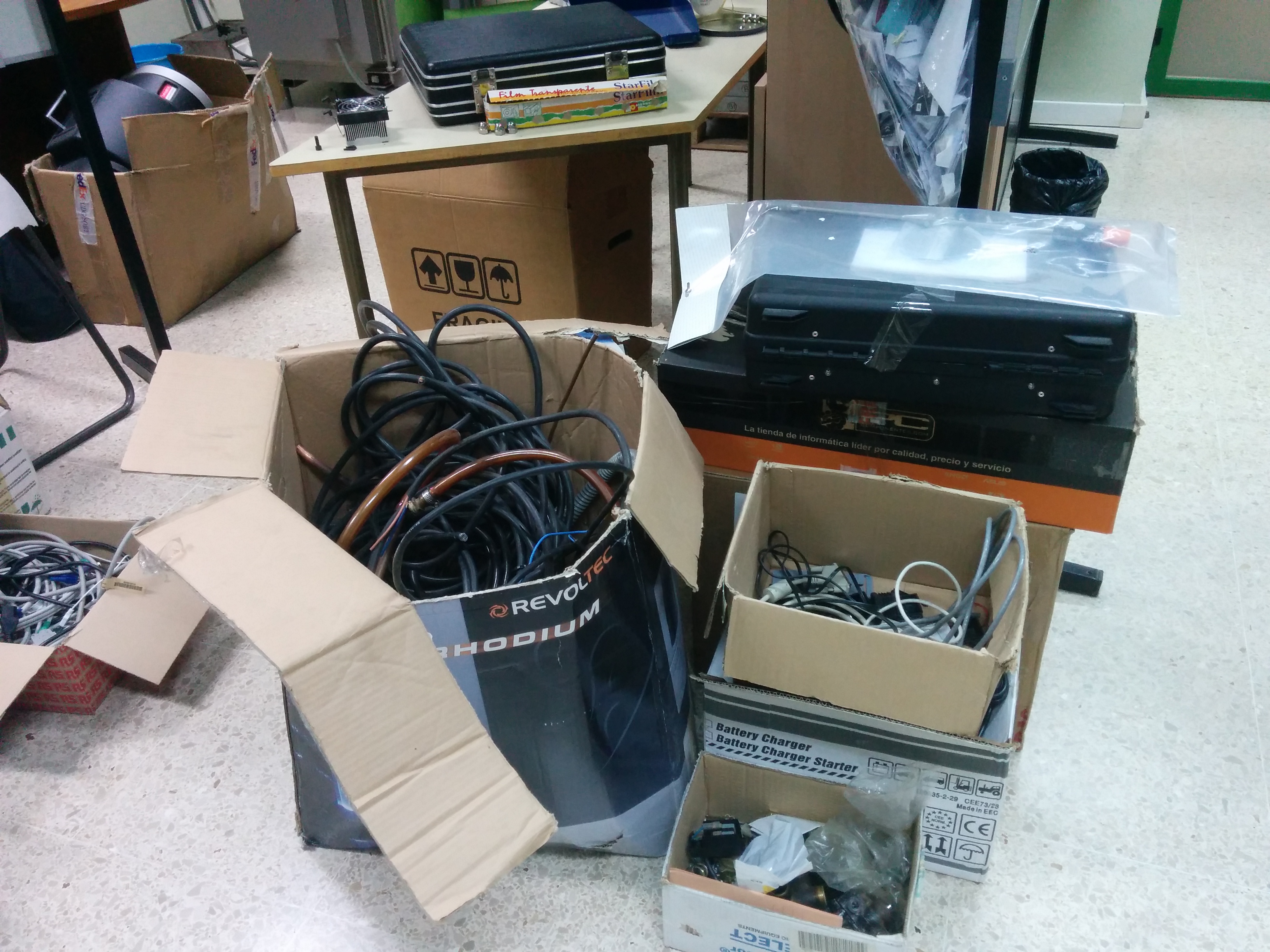}
\end{subfigure}
\end{subfigure}
\begin{subfigure}{0.84\textwidth}
\centering
\includegraphics[width=\textwidth]{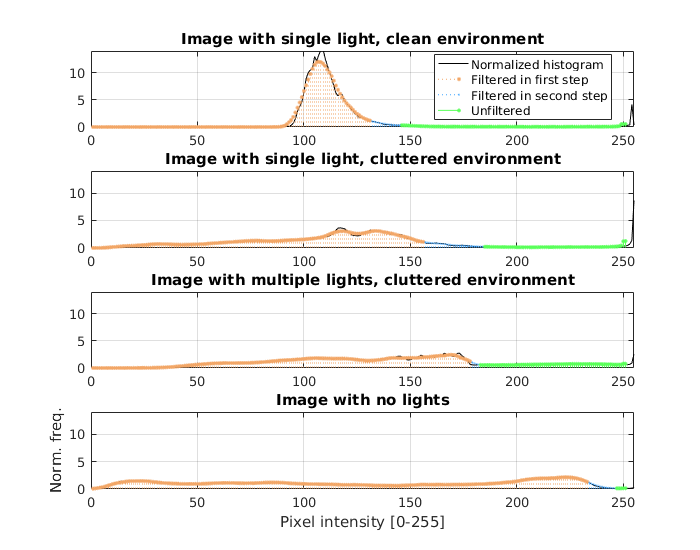}
\end{subfigure}
\caption{Normalized and smoothed histogram and threshold values for multiple images with different contents.}
\label{fig:hist}
\end{figure*}

Figure~\ref{fig:hist} shows several examples of the normalized histogram of different images under different conditions. As can be inferred, histograms of images with lamps usually have three main regions: (i) the low and medium spectrum that contains most of the values in the image, corresponding to the dark background elements; (ii) a fairly flat region of high-intensity values after a peak, corresponding to bright objects and contours of light sources; and (iii) a peak in the highest intensity region, corresponding to saturated values, usually centres of light sources and reflections.

Light surfaces are usually contained in regions (ii) and (iii), while values in region (i) should be filtered. First, we apply a smoothing on the normalized histogram using a moving average of $11$ samples. Then, we use the smoothed histogram $\hat{\mathcal{F}}[k]$ to obtain a threshold:
\begin{align}
\mathcal{T}_L &= \max_{}\{k \mid \hat{\mathcal{F}}[k] < B_u\} .
\end{align}

\begin{figure*}[h]
\centering
\begin{subfigure}{0.30\textwidth}
\centering
\includegraphics[width=\textwidth]{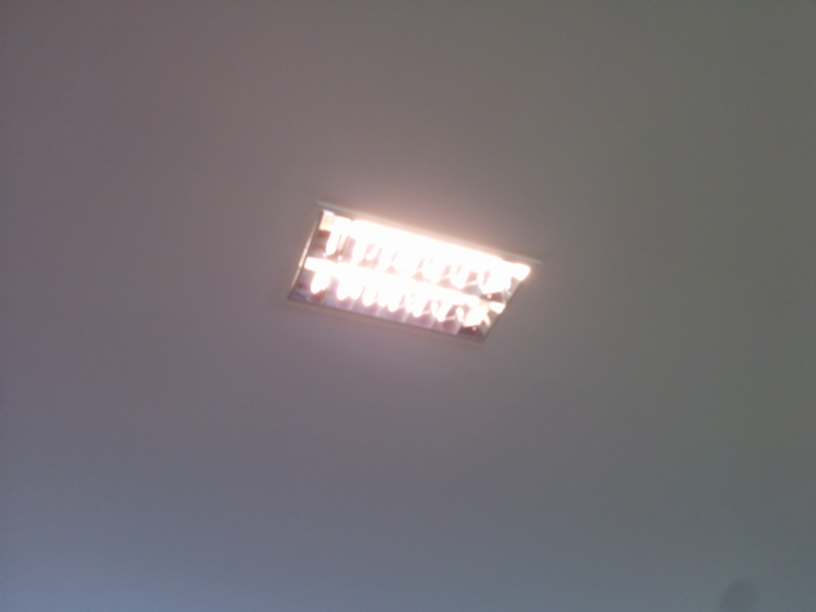}
\caption{Original image}
\end{subfigure}
\begin{subfigure}{0.30\textwidth}
\centering
\includegraphics[width=\textwidth]{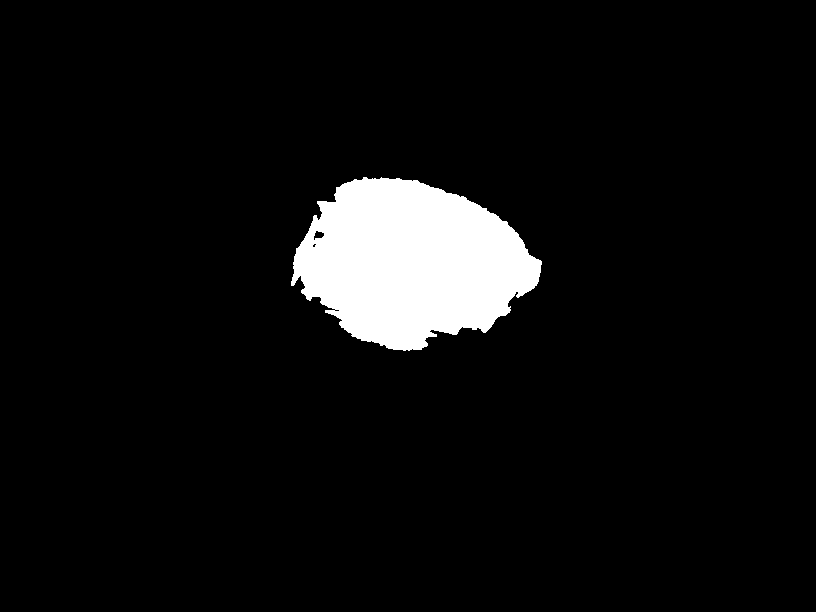}
\caption{Threshold with $\mathcal{T}_L$}
\label{subfig:thl}
\end{subfigure}
\begin{subfigure}{0.30\textwidth}
\centering
\includegraphics[width=\textwidth]{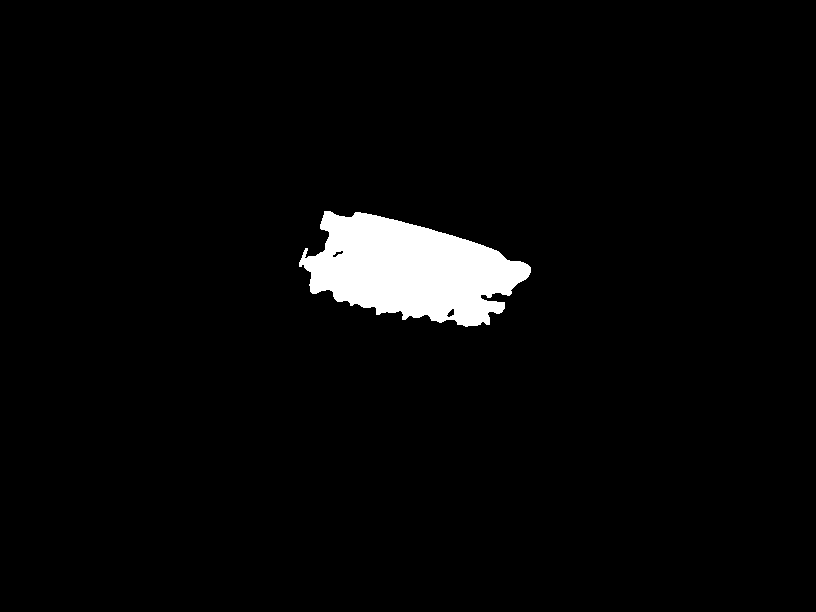}
\caption{Threshold with $\mathcal{T}_U$}
\label{subfig:thu}
\end{subfigure}
\caption{Binary images resulting from the threshold operation with different values.}
\label{fig:th}
\end{figure*}

Using $\mathcal{T}_L$ as the intensity threshold, the background is mainly removed, but the resulting binary image still contains bright areas surrounding the main light source, as shown in Figure~\ref{subfig:thl}. Therefore, we apply another step to filter the remaining noise:
\begin{align}
\mathcal{T}_U &= \min_{\mathcal{T}_L \leq k}\{k \mid \mathrm{d}\hat{\mathcal{F}}[k] > B_l\} ,
\end{align}
which yields blobs with outlines much closer to the light surface, as shown in Figure~\ref{subfig:thu}. We use $B_u=1$ and $B_l=0.015$ in our experiments.

\subsubsection{Shape estimation}

We assume that the lamps to be detected have an effective light surface with a flat polygonal shape. Therefore, to obtain estimations of the lamp pose based on a bright blob, we apply the method described in Algorithm~\ref{alg:bpoly} with different values of $k$. Thus, we can get estimations for lamps with different number of sides, and even generalize this method to circular shapes if we consider them like polygons with a sufficiently large number of sides. Nevertheless, all the tested models have a rectangular light surface, so we fix $k=4$ in the experiments.

For the sake of simplicity, we relax the notation to imply that all the indexing is done in a circular manner, so $X_{i} = X_{1 + ((i-1)\mod n)}$, with $n$ the length of $X$. The score function $\mathcal{S}(P, i)$ defines the cost that the algorithm will try to minimize, while the function $\mathcal{G}(P, i)$ generates a new point for the current iteration.

\begin{algorithm}[h]
\caption{Fit polygon}
\label{alg:bpoly}
\begin{algorithmic}[1]
\Require
\Statex $P$ is a sequence of points
\Statex $k$ is the number of sides in the resulting polygon
\Ensure
\Statex $C$ is a sequence of $k$ points representing an approximated polygon for $P$
\Statex
\State $C \gets $\Call{ConvexHull}{$P$}
\State $n \gets \mathbf{len}~C$
\State $S \gets [0,\ldots,0]_{n}$
\For{$i \gets 1, n$}
\State $S_{i} \gets \mathcal{S}(C,i)$
\EndFor
\While{$n > k$}
\State $m \gets \arg\,\min_{1 \leq i \leq n}(S_i)$
\State $C_m \gets \mathcal{G}(C, m)$
\State \Call{RemoveElement}{$C$, $m+1$}
\State \Call{RemoveElement}{$S$, $m+1$}
\State $S_{m} \gets \mathcal{S}(C,m)$
\State $n \gets n-1$
\EndWhile
\end{algorithmic}
\end{algorithm}

The algorithm can produce a new polygon that is strictly inner or outer to the original one. Let $\mathcal{A}(a,b,c)$ be the area of the triangle defined by the points $\{a, b, c\}$ and $\mathcal{I}(P, i)$ be the intersection point of the lines defined by $\{P_{i-1}, P_i\}$ and $\{P_{i+1}, P_{i+2}\}$. If we want to minimize the area of the resulting polygon, the functions may be defined as
\begin{align}\label{eq:innerAS}
\mathcal{S}_{inner}(P, i) =& \mathcal{A}(P_i, P_{i+1}, P_{i+2}), \\\label{eq:innerAG}
\mathcal{G}_{inner}(P, i) =& P_i
\end{align}
for an inner polygon, and
\begin{align}\label{eq:outerAS}
\mathcal{S}_{outer}(P, i) =& \mathcal{A}(P_i, \mathcal{I}(P, i), P_{i+1}), \\\label{eq:outerAG}
\mathcal{G}_{outer}(P, i) =& \mathcal{I}(P, i)
\end{align}
for an outer polygon.
Analogous reasoning can be applied if the perimeter is to be minimized, but we use the area as the cost function since it yielded slightly better results in our experiments.

Given that the resulting polygon does not need to be strictly inner or strictly outer, we use a combination of equations~\ref{eq:innerAS}, \ref{eq:innerAG}, \ref{eq:outerAS} and \ref{eq:outerAG}, generating two scores for each point, which improves the results.

\subsubsection{Pose estimation and filtering}

Once a set of four 2D points is obtained, we have to find an object pose that matches these to the 3D points of the lamp model. This task is performed by solving a PnP ({\em Perspective-and-Point}) problem using an iterative method based on Levenberg-Marquardt optimization~\cite{Levenberg,Marquardt} to find a pose that minimizes reprojection error, that is the sum of squared distances between the observed projections and the projected 3D points.

A set of simple filters can be applied at this point to easily discard false positives as soon as possible, avoiding complex computations on regions that would not yield any detection.

\begin{figure}[h]
\centering
\includegraphics[width=0.7\linewidth]{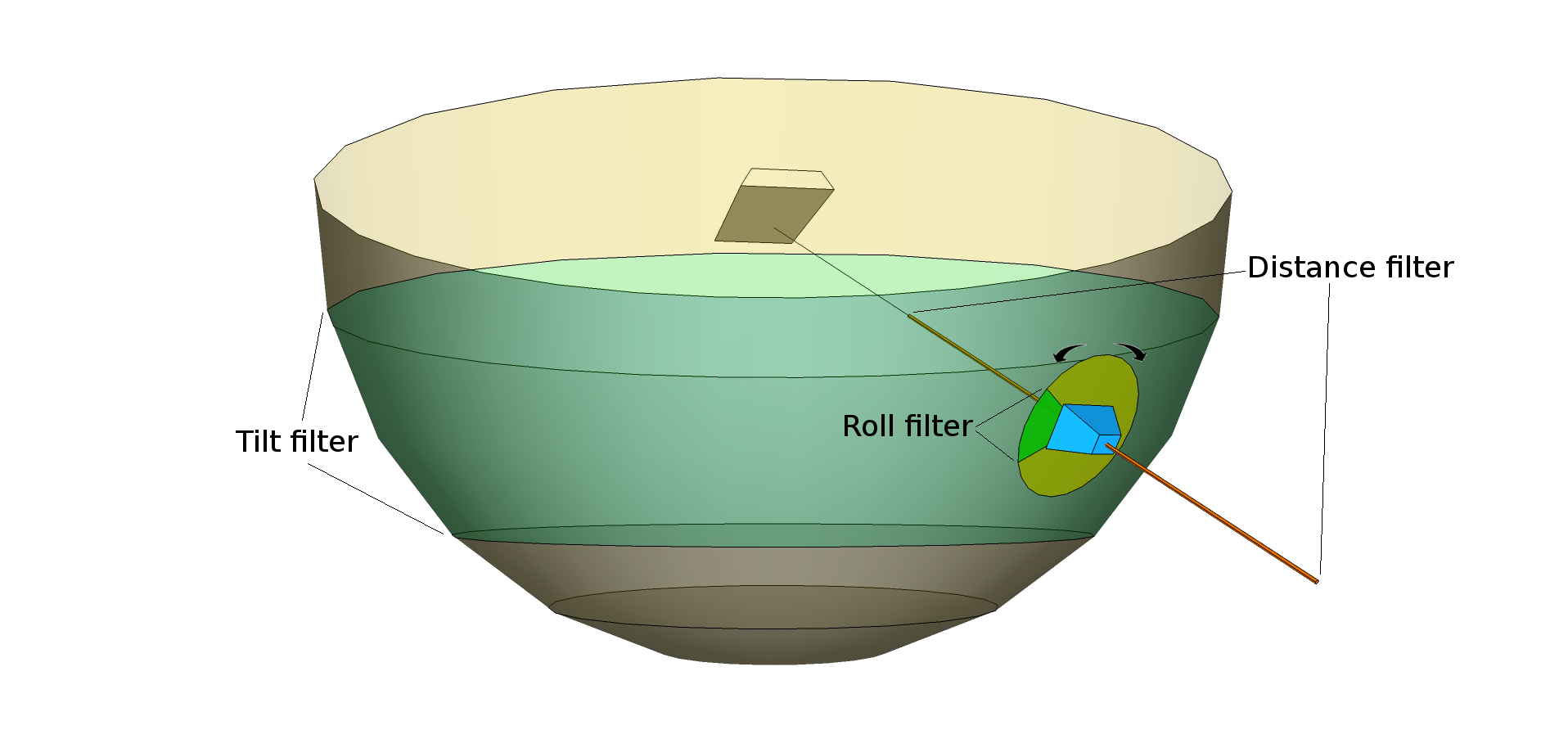}
\caption{Outline of the pose filters used to discard false positives based on the supposed camera orientation and distance range.}
\label{fig:filters}
\end{figure}

We use three tuneable thresholds based on the usual lamp positions in building interiors to discard erroneous poses, based on distance, roll and tilt between the supposed lamp and the camera. Figure \ref{fig:filters} shows these thresholds and the corresponding valid regions for the detected lamps with respect to the camera. The thresholds can be adjusted depending on how accurately we can determine the orientation of the camera. If no information is available, a good compromise is to set the roll threshold to $\pm 20^\circ$ and the tilt threshold to $[10^\circ, 90^\circ]$ from the ceiling plane.

\subsection{Pose and model selection}
\label{sect:poseModelSelection}

For each candidate pose, we select all the templates that correspond to a pose that is close in position and orientation for all lamp models in the database. The projected visible edges from the model templates have to be compared with the edge information retrieved from the image. For this purpose, we use the Fast Directional Chamfer Matching (FDCM)~\cite{Liu}.

\begin{figure*}[h]
\centering
\begin{subfigure}{0.235\textwidth}
\centering
\includegraphics[width=\textwidth]{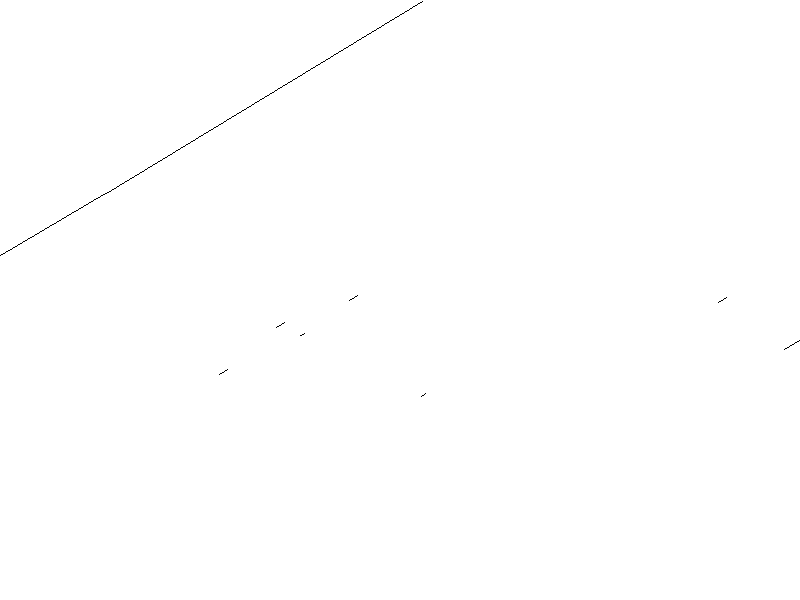}
\caption{Edges}
\end{subfigure}
\begin{subfigure}{0.235\textwidth}
\centering
\includegraphics[width=\textwidth]{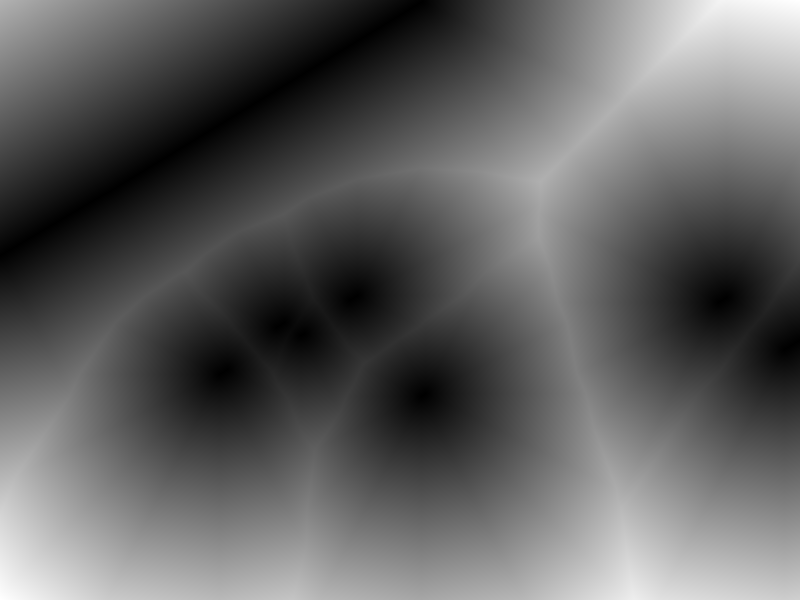}
\caption{Dst. transform}
\end{subfigure}
\begin{subfigure}{0.235\textwidth}
\centering
\includegraphics[width=\textwidth]{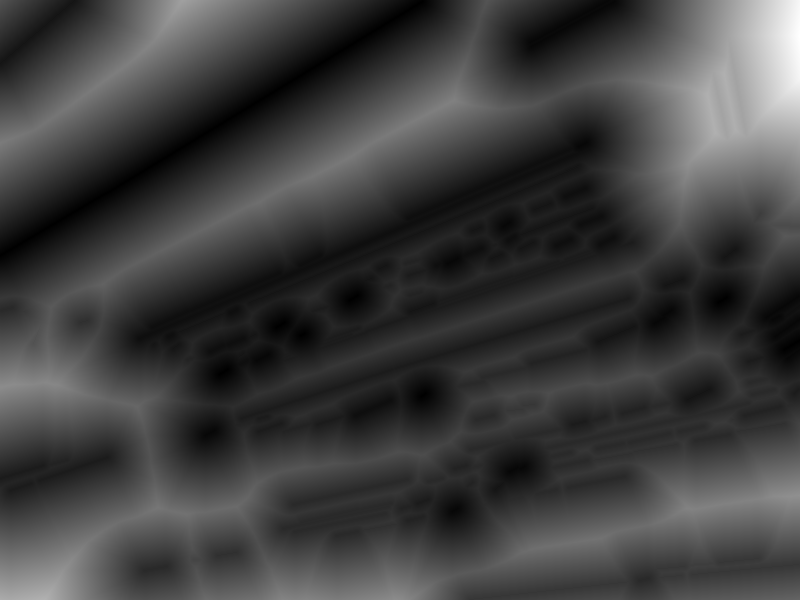}
\caption{DT3$_V$}
\end{subfigure}
\begin{subfigure}{0.235\textwidth}
\centering
\includegraphics[width=\textwidth]{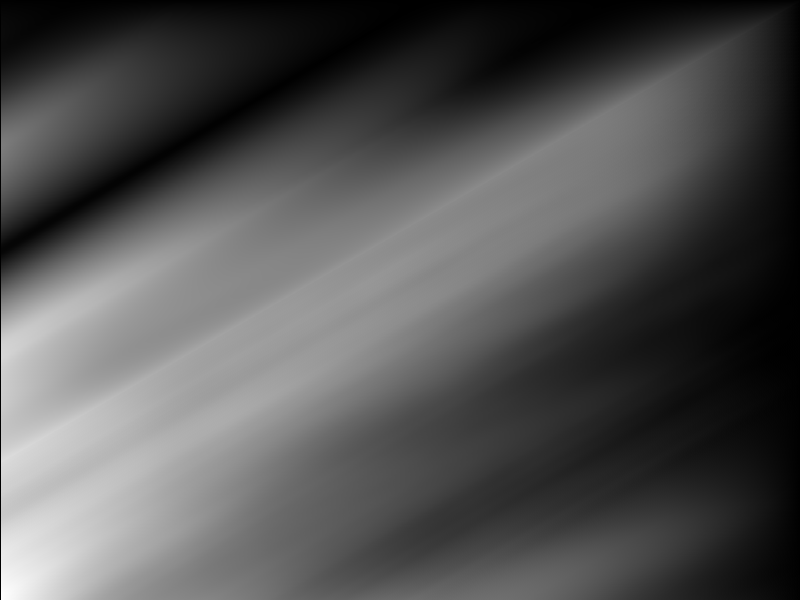}
\caption{IDT3$_V$}
\end{subfigure}
\caption{Output for an orientation image in different stages of the construction of the IDT3$_V$, for the image in Figure \ref{subfig:det_a2}.}
\label{fig:idt3v}
\end{figure*}

First, we determine a region of interest (ROI) that contains all the bounding boxes of the possible templates with a small extra margin. This process is very fast, as the bounding boxes can be precomputed with the templates. Then, for this region, we compute the three-dimensional distance transform representation, DT3$_V$, and the integral distance transform, IDT3$_V$, needed for the matching. This procedure involves several sub-steps, as described in~\cite{Liu}: (i) we use the Line Segment Detector \cite{lsd} to extract edge information in the ROI; (ii) we divide the set of edges depending on their quantized direction in 60 bins and draw each of them in the corresponding binary image; (iii) we perform a distance transform on each of the binary images; (iv) we update the images with a forward and a backward recursion to account for the distance between the different orientations with a weighting factor $\lambda = 100$, obtaining the DT3$_V$; and (v) for each image, we compute the cumulative sum of its elements in the direction corresponding to the image orientation, obtaining the IDT3$_V$. Figure \ref{fig:idt3v} shows the output of one of the orientation images at different stages of the process.

The use of the IDT3$_V$ and the search optimizations presented by Liu \emph{et al.} allow for a very fast evaluation of templates, with empirical evidence of sublinear complexity~{\cite{Liu}}. We evaluate all the selected templates for every model in the database and keep the one with the lowest distance. Having this done, we can identify the lamp model and refine the pose.

\subsection{Pose refinement and scoring}
\label{sect:poseRefinementScoring}

With the aim of providing adequate scoring for the detected poses and refining their characteristics, we use the proposal of Imperoli and Pretto~\cite{D2CO}, i.e., the Direct Directional Chamfer Optimization (D\textsuperscript{2}CO).
The scoring function consists of an average of the cosine similarity between the normal direction of each projected raster point and the local gradient direction for the corresponding image point. Therefore, for a pose $\mathbf{\Omega}$, the score is given by
\begin{align}
\phi(\mathbf{\Omega}) = \frac{1}{n}\sum_{i=1}^n |\cos(\bm g(\bm p_i)) - \bm n(\bm p_i)|,
\end{align}
where $\bm p_i$ is the $i$-th projected raster point; $\bm g(\bm p_i)$ the local gradient direction for the image point and $\bm n(\bm p_i)$ the normal direction of $\bm p_i$.

\subsection{3D pose processing}
\label{sect:3DPoseProcessing}

Having an image $I$ as input for the detection algorithm, its output consists of a set of weighted poses:
\begin{align}
S_I = \left\{\left(\mathbf{\Omega}_O^{(1)}, \pi^{(1)}\right),\hdots,
\left(\mathbf{\Omega}_O^{(N_I)}, \pi^{(N_I)}\right)\right\} ,
\end{align}
where $\mathbf{\Omega}_O^{(n)} \in \mathbb{SE}(3)$ is the relative object-camera pose and $\pi^{(n)}$ is the score for the $n$-th surviving detection of a lamp in the image.

\begin{figure}[h]
\centering
\includegraphics[width=0.8\linewidth]{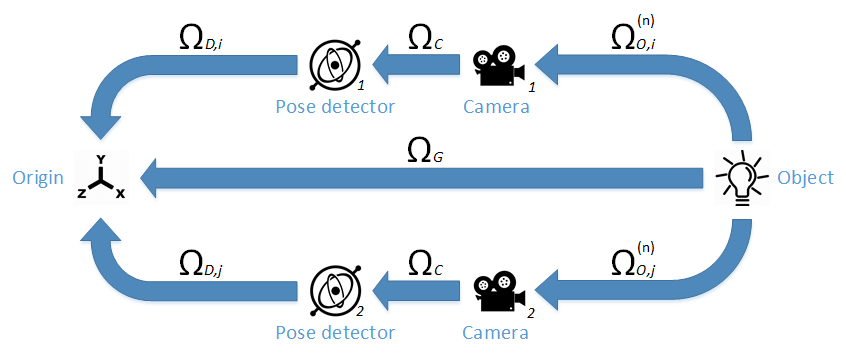}
\caption{Pose transformation scheme.}
\label{fig:3d}
\end{figure}

To correctly translate these relative poses to a common origin that allows the comparison of outputs from different images or frames in a video, we need a pose detector that registers a pose $\mathbf{\Omega}_D$ for each video frame. Therefore, to obtain a global pose $\mathbf{\Omega}_G$ relative to the predefined origin, we need to apply the transformation shown in Figure~\ref{fig:3d} for a frame $i$, combining: (i) the relative object-camera pose obtained from the detection algorithm; (ii) the pose $\mathbf{\Omega}_C$ representing a rigid transformation between the detector and the camera; and (iii) the pose registered by the detector with respect to the origin:
\begin{align}
\mathbf{\Omega}_G = \mathbf{\Omega}_{D,i} \, \mathbf{\Omega}_C \, \mathbf{\Omega}_{O,i} .
\end{align}

Alternately, with the aim of restoring the object pose for a different camera position in another frame $j$, the inverse transformation is applied:
\begin{align}
\mathbf{\Omega}_{O,j} = \mathbf{\Omega}_C^{-1} \, \mathbf{\Omega}_{D,j}^{-1} \, \mathbf{\Omega}_G .
\end{align}

When processing video sequences, it is possible to obtain more than one detection for the same lamp, so a mechanism to avoid these duplicates should be included that simultaneously leverages this combined information. Therefore, if $S_k = \{(\mathbf{\Omega}_{G,k}^{(n)}, \pi_k^{(n)})\}, n \in 1,\hdots,N_k$ is the set of detected poses and scores for the $k$-th object found, the resulting pose of this object is computed based on a weighted average of each one of its detections:
\begin{align}
\overline{\mathbf{\Omega}}_k = \sum_{n=1}^{N_k} \pi_k^{(n)} \mathbf{\Omega}_{G,k}^{(n)} .
\end{align}

Let $\mathbf{R} \in \mathbb{SO}(3)$ be the orientation part of a pose $\bf\Omega$. The arithmetic mean of a set of orientations $\overline{\mathbf{R}} = \sum_{n=1}^{N} \mathbf{R}^{(n)}$ does not necessary lie on $\mathbb{SO}(3)$, so we need to apply a different method to perform the average. Moakher~\cite{Moakher} demonstrated that a valid average can be calculated via the orthogonal projection of $\overline{\mathbf{R}}$ as follows:
\begin{align}
\overline{\mathbf{R}}' = \begin{cases}
\mathbf{V}\mathbf{U}^T & \text{if } \det(\overline{\mathbf{R}}^T) > 0 \\
\mathbf{V}\mathbf{H}\mathbf{U}^T & \text{otherwise} ,
\end{cases}
\end{align}
where $\mathbf{U}$ and $\mathbf{V}$ are estimated from the singular decomposition value of $\overline{\mathbf{R}}^T =~\mathbf{U}\mathbf{\Sigma}\mathbf{V}^T$, and $\mathbf{H} = \text{diag}(1, 1, -1)$.

For each new detection we have to decide if it comes from a new lamp or from a previously detected one. For this, we use a simple decision procedure: if the Euclidean distance between the new detection and the closest previously detected object is smaller than a predefined threshold, the new detection is added to the lamp registry; otherwise, the detection is considered to be from a new lamp.

\subsection{Automatic inclusion of lighting information in BIM}
\label{sect:automaticInclusionLighting}

Given the current model of the building with a set of thermal zones and their associated geometric information, we can automatically include each one of the detected lamps in the appropriate thermal zone by means of their absolute position with respect to a known origin. If this origin is correctly located in the building model coordinates, a simple transformation is the only thing required to map the detections in the building model world; then, we perform a point-in-polyhedron test for each thermal zone to identify the one that contains the lamp.

\begin{figure*}[h]
\centering
\begin{subfigure}{0.36\textwidth}
\centering
\includegraphics[width=\textwidth]{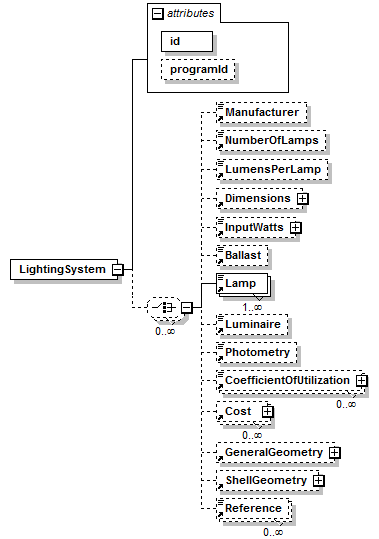}
\end{subfigure}
\begin{subfigure}{0.63\textwidth}
\centering
\includegraphics[width=\textwidth]{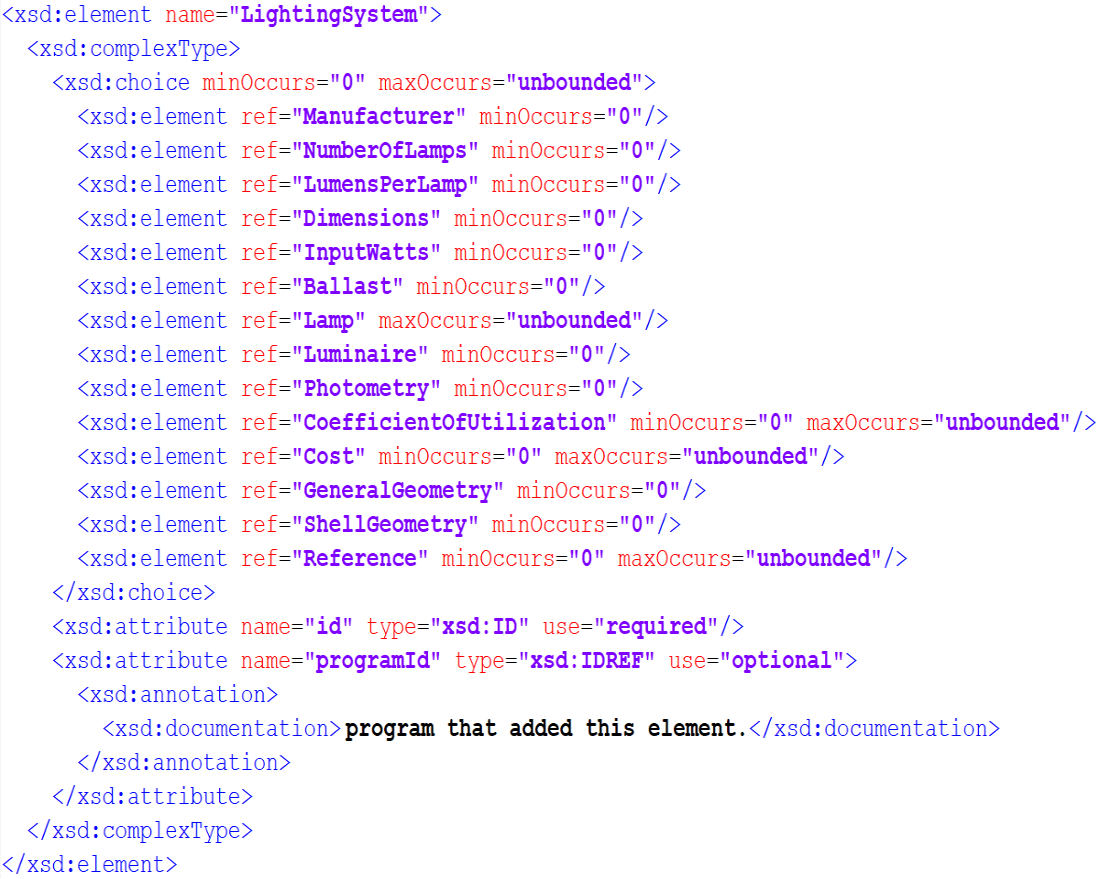}
\label{subfig:det_a}
\end{subfigure}
\caption{Green Building XML schema definition for the \texttt{LightingSystem} element, which includes the lighting information that can be automatically generated by the detection system.}
\label{fig:gbxml}
\end{figure*}

Green Building XML (gbXML)~\cite{gbxml} is an open schema created to facilitate the transference of building data stored in Building Information Modeling (BIM) to engineering analysis tools. We use this open standard to demonstrate the automatic inclusion of the information gathered with the detection algorithm in the BIM of the building. Figure \ref{fig:gbxml} shows part of the gbXML used to define the lighting information.

The fields \texttt{Lamp}, \texttt{NumberOfLamps}, and \texttt{ShellGeometry} can be automatically filled. Then, depending on the available data of the lamp model present in the database, information for other fields can also be included, such as \texttt{Luminaire} and \texttt{Cost}. The \texttt{LightingSystem} element is later referenced in the corresponding \texttt{Space} via its \texttt{Lighting} element with the \texttt{lightingSystemIdRef} attribute.

\section{Results}
\label{sec:results}

\begin{figure}[h]
\centering
\begin{subfigure}{0.32\linewidth}
\centering
\includegraphics[width=0.8\textwidth]{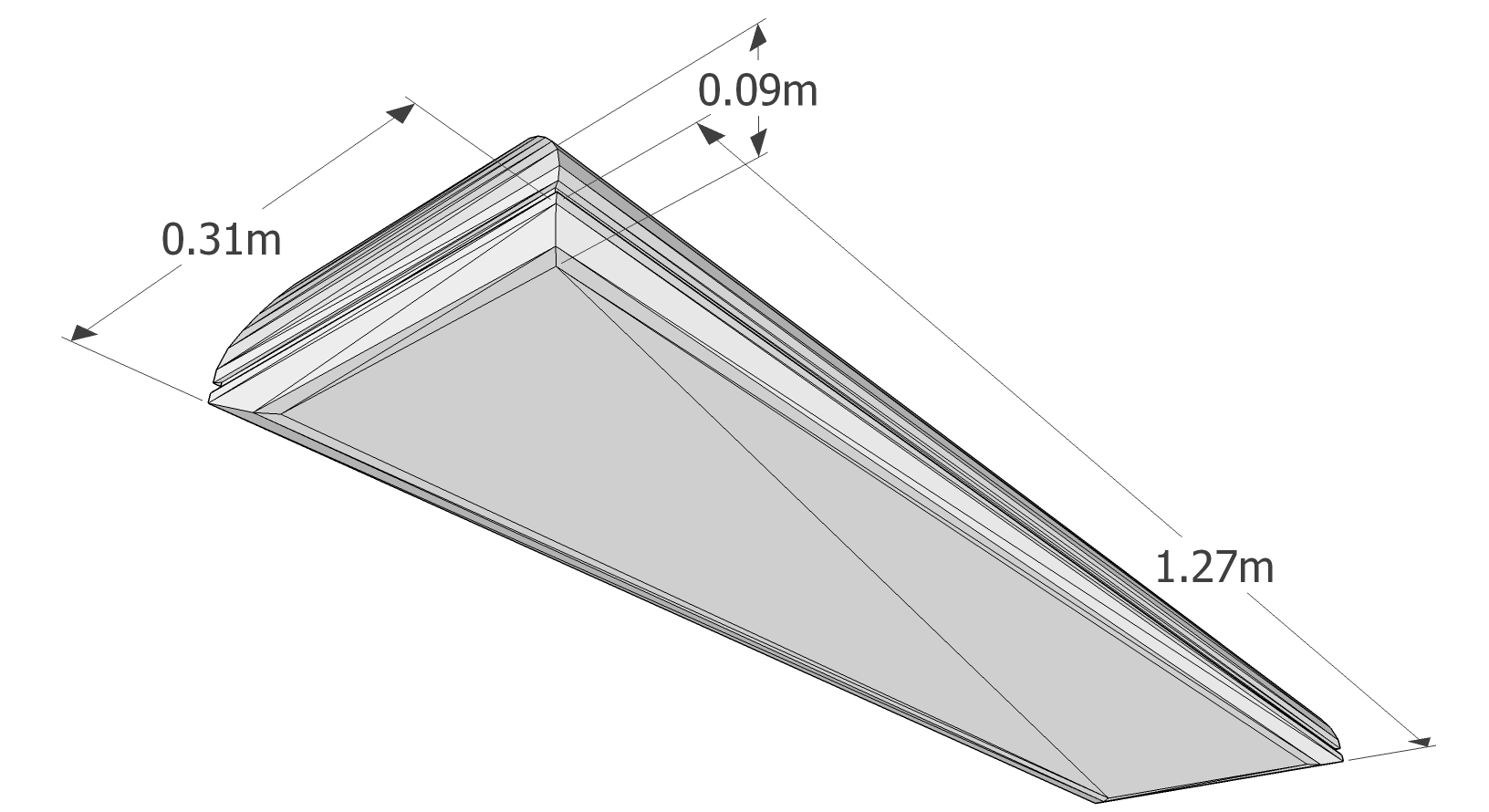}
\caption{Model 1}
\end{subfigure}
\begin{subfigure}{0.32\linewidth}
\centering
\includegraphics[width=0.8\textwidth]{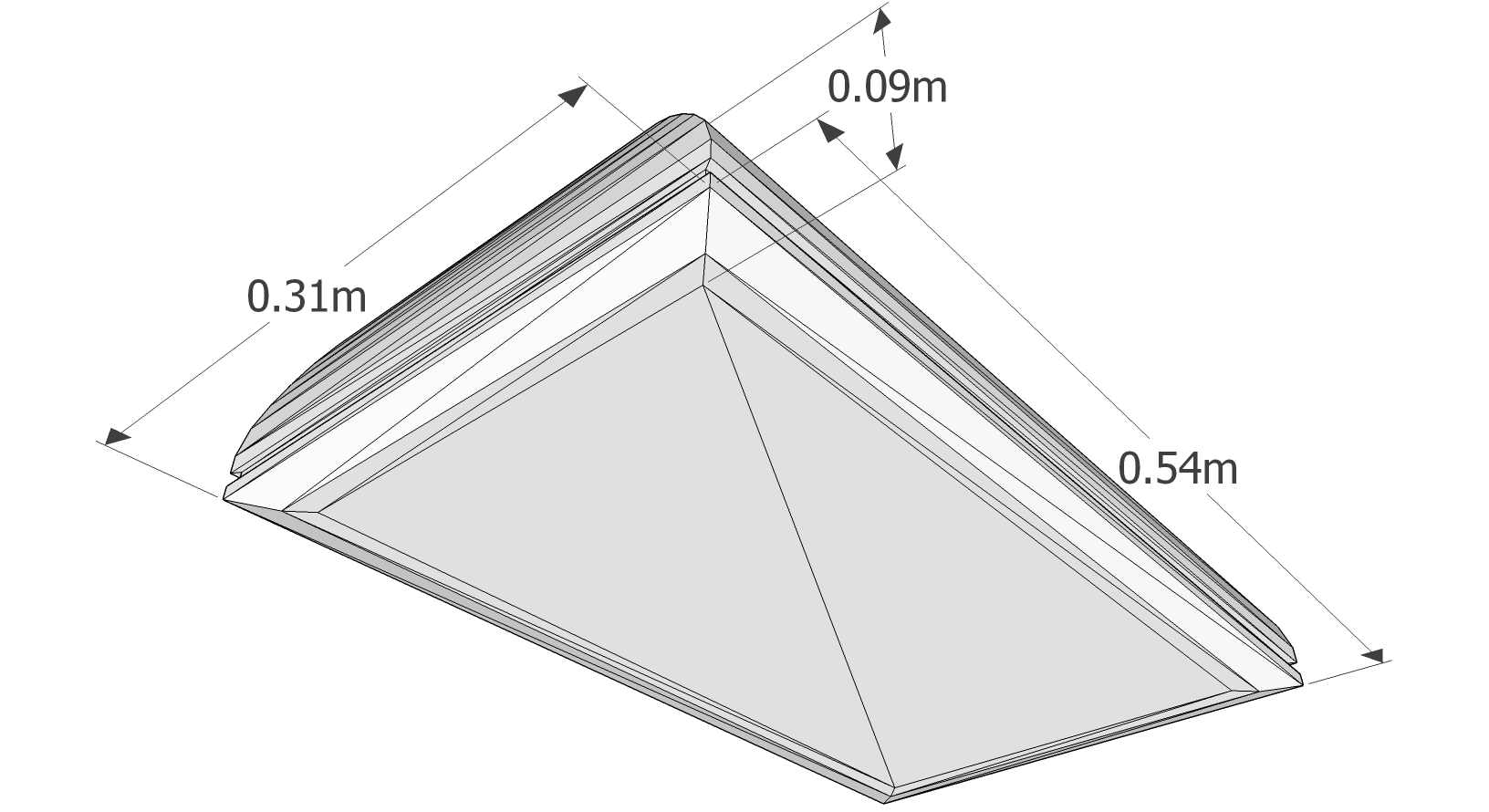}
\caption{Model 2}
\end{subfigure}
\begin{subfigure}{0.32\linewidth}
\centering
\includegraphics[width=0.8\textwidth]{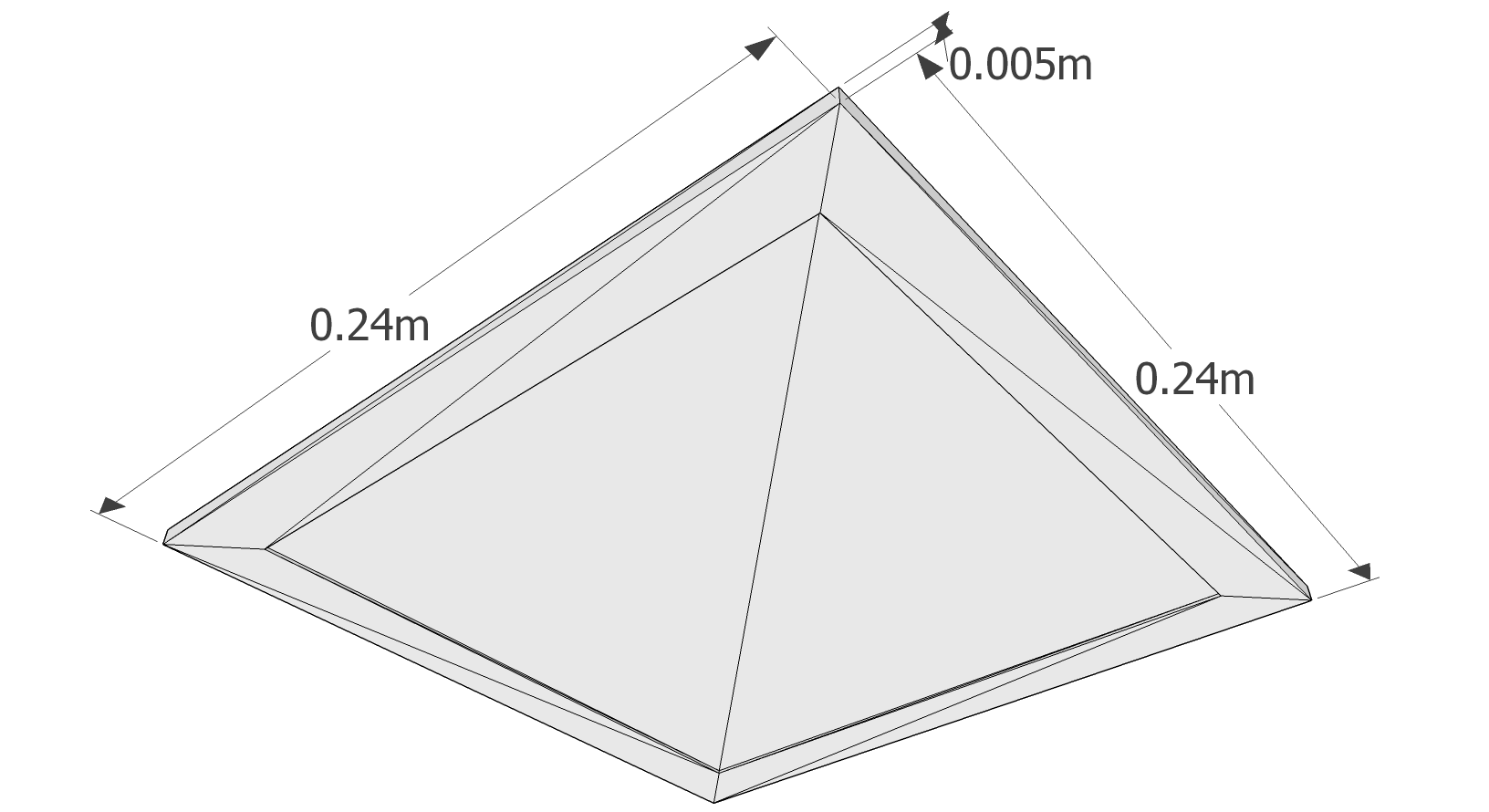}
\caption{Model 3}
\end{subfigure}
\caption{Lamp models used for the experiments.}
\label{fig:lamps}
\end{figure}

Our methodology has been tested for the three lamp models presented in Figure \ref{fig:lamps}, with their basic characteristics in Table~\ref{tab:bulbs}. The main tests of the algorithm, described in Section \ref{subsec:images}, were performed on individual images. Moreover, we tested the complete system on a video sequence, generating the BIM data of the lamps for the corresponding thermal zones in the building model; this is explained in Section  \ref{subsec:video}. We used the OpenCV library~\cite{OpenCV} for the implementation of the proposed method.

\begin{table}[h]
\centering
\small
\caption{Characteristics of the fluorescent bulbs in the lamp models.}
\begin{tabular}{|c|c|c|c|c|c|c|c|}
\hline
\textbf{Model} & \textbf{No.} & \textbf{Brand} & \textbf{Series} & \textbf{Tech.} & \textbf{Power} & \textbf{Brightness} & \textbf{Color} \\
\hline
1 & 2 & Osram & Lumilux & Fluor. & 36 W & 3350 lm & 4000 K \\
2 & 2 & Sylvania & Lynx & Fluor. & 36 W & 2800 lm & 3000 K \\
3 & 2 & Osram & Dulux & Fluor. & 26 W & 1800 lm & 4000 K \\
\hline
\end{tabular}
\label{tab:bulbs}
\end{table}

\subsection{Individual images}
\label{subsec:images}

A total of 51 images were used for our tests, with a resolution of $800 \times 600$ pixels and without applying any correction filter. The images are divided in three categories, as shown in Table \ref{tab:dataset}, included a brief description of the lamp positioning and environment.

\begin{table}[h]
\centering
\small
\caption{Categories of the dataset of images used in the experiments.}
\begin{tabular}{|p{6.8cm}|c|c|c|}
\hline
\textbf{Area} & \textbf{Lamp} & \textbf{No. imgs.} & \textbf{No. dets.} \\
\hline
Laboratory, two rooms, lamps suspended 50 cm from the ceiling, only two external windows, 1 m away from the closest lamps & Model 1 & 22 & 57 \\
\hline
Hallway, lamps suspended 40 cm from the ceiling, external windows at one side & Model 2 & 23 & 25 \\
\hline
Reception, big open area, second floor, lamps fixed at the ceiling, bright environment & Model 3 &  6 & 14 \\
\hline
\textbf{TOTAL} & & \textbf{51} & \textbf{96} \\
\hline
\end{tabular}
\label{tab:dataset}
\end{table}

\begin{figure}[b!]
\centering
\includegraphics[width=0.35\linewidth]{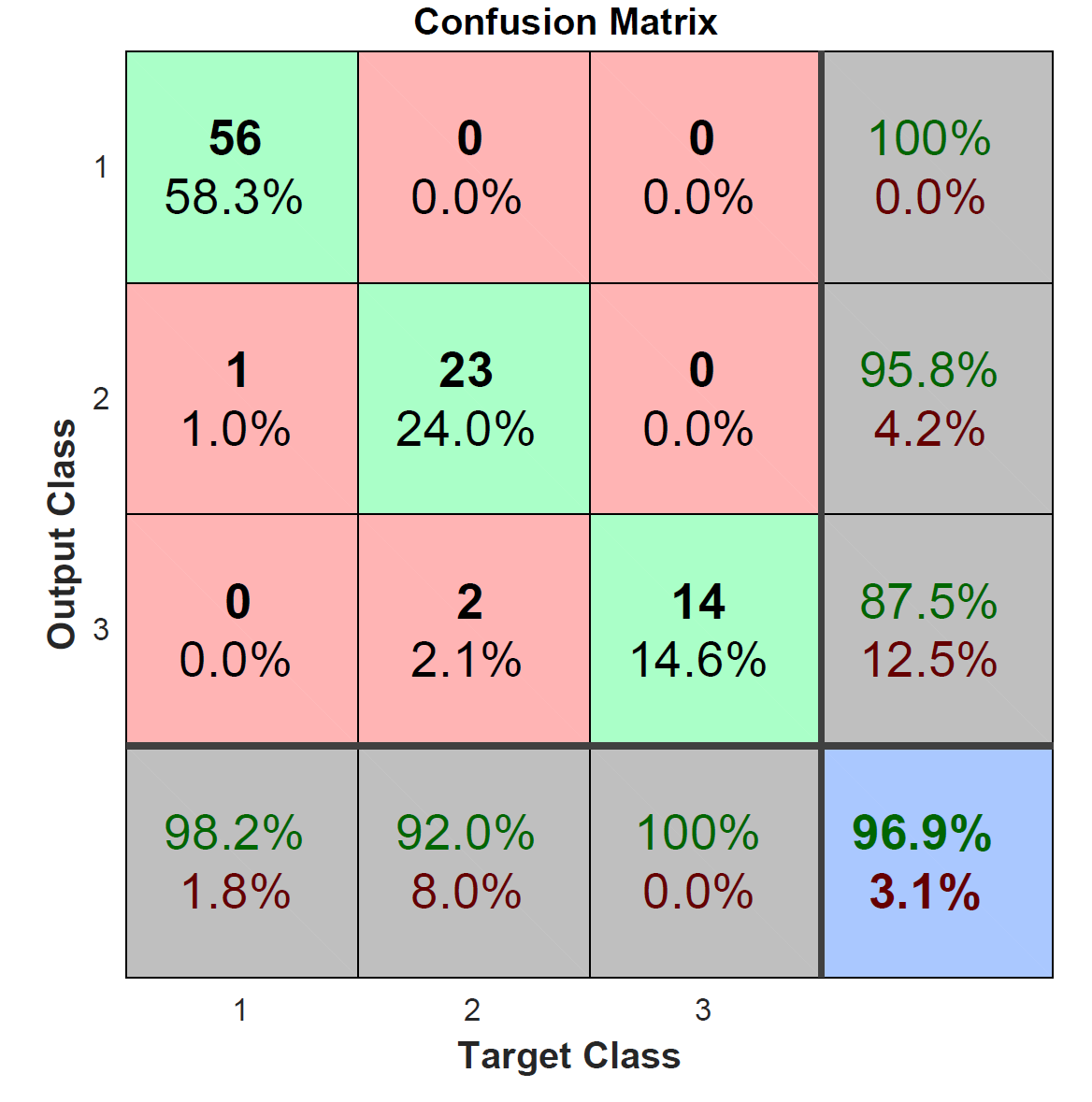}
\caption{Confusion matrix for the images used in the experiments.}
\label{fig:confusion}
\end{figure}

\begin{figure}[b!]
\centering
\includegraphics[width=0.5\linewidth]{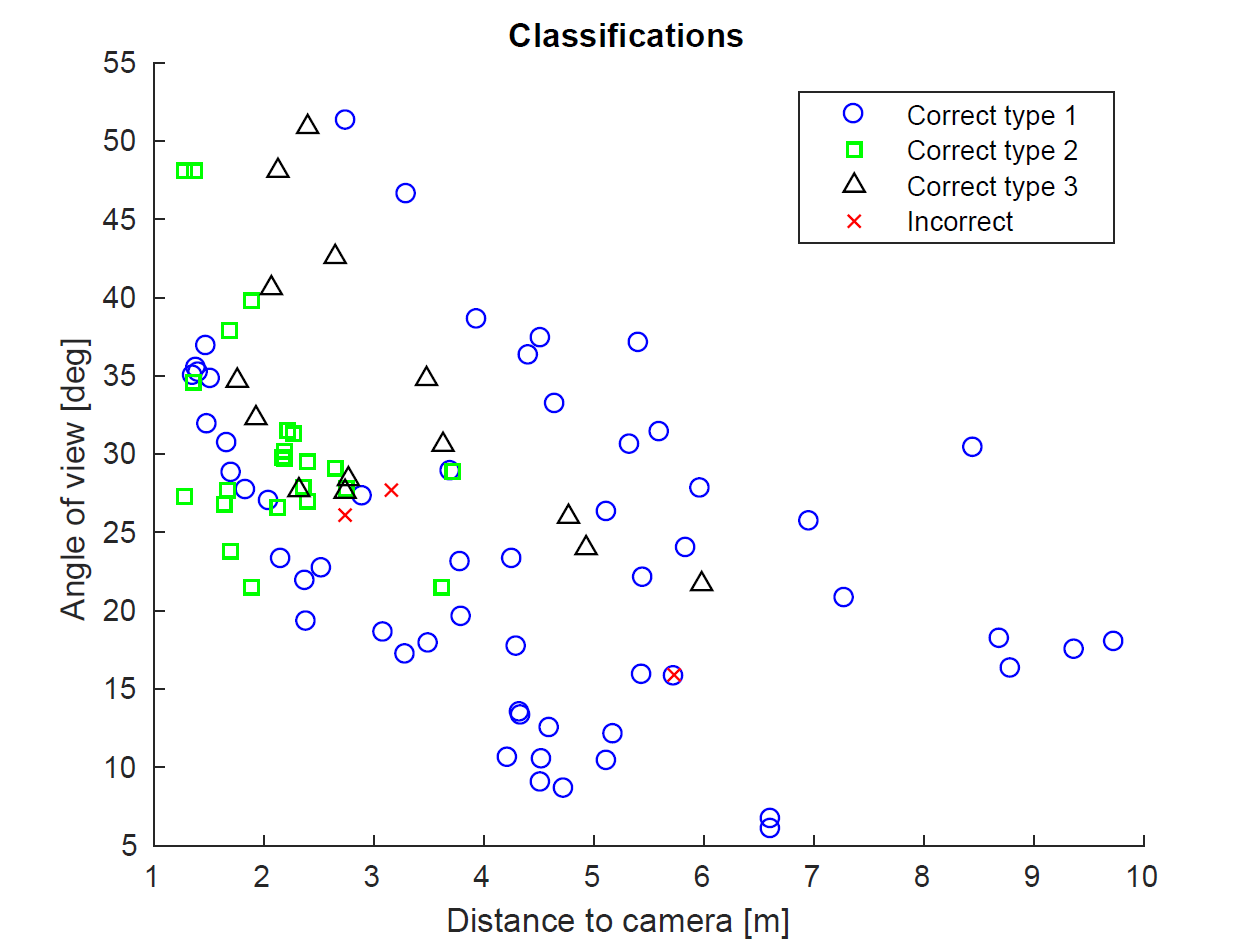}
\caption{Scatter plot of detections for the three classes with respect to distance to camera and angle of view (to ceiling plane).}
\label{fig:scatter_da}
\end{figure}

\begin{figure}[h]
\centering
\includegraphics[width=0.5\linewidth]{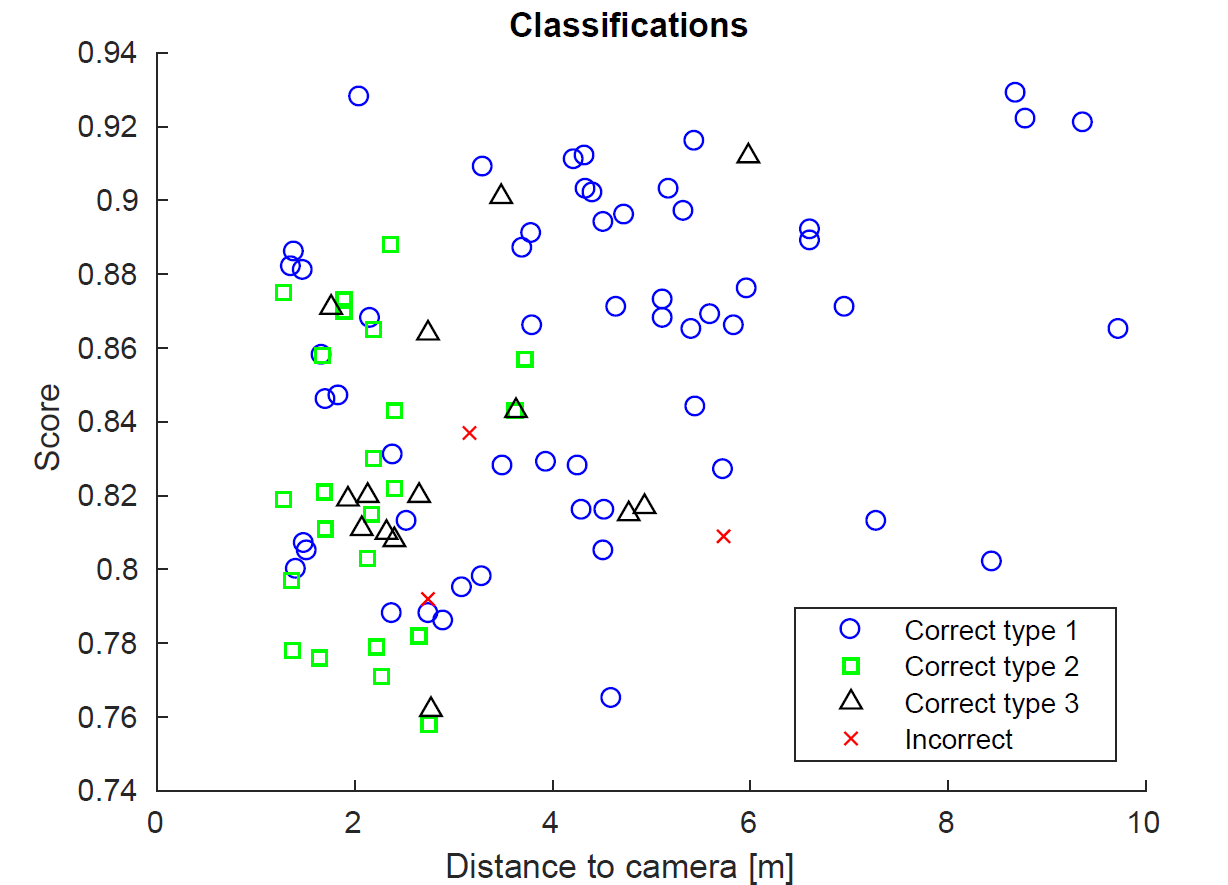}
\caption{Scatter plot of detections for the three classes with respect to distance to camera and score.}
\label{fig:scatter_ds}
\end{figure}

\begin{figure}[h!]
\centering
\includegraphics[width=0.5\linewidth]{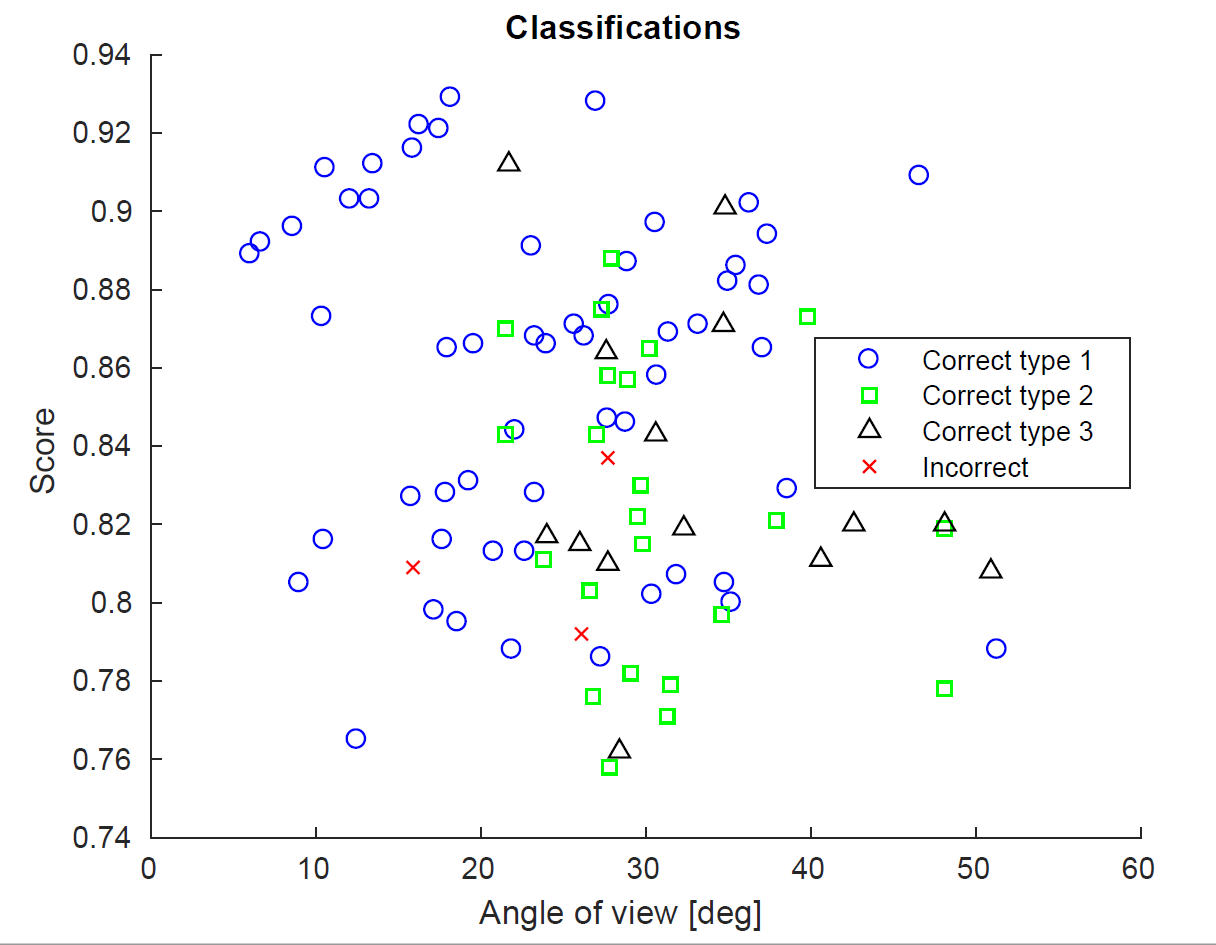}
\caption{Scatter plot of detections for the three classes with respect to angle of view (to ceiling plane) and score.}
\label{fig:scatter_as}
\end{figure}

Figure~\ref{fig:confusion} presents the confusion matrix of the detections in the images for the three classes of lamps (models 1, 2 and 3), showing that 96.9\% of the 96 detected lamps were correctly identified. We also include some scatter plots of classifications with respect to distance and angle, distance and score, and angle and score in Figures \ref{fig:scatter_da}, \ref{fig:scatter_ds} and \ref{fig:scatter_as}, respectively.

There were three missclassifications: one lamp of model 1 was missclassified as model 2 while two model 2 lamps were missclassified as model 3. In the first case, the fact that model 1 and 2 have a similar shape caused that the edge information of model 2 matched the detection better in this case. In the second case, the incorrect classifications are due to the the fact that model 3 has a very thin frame that can sometimes match the light surface of bigger models better than the entire frame of the correct model. These problems arise in poor background and lighting conditions that causes the extracted edge information to be inaccurate or vague. Figure \ref{fig:scatter_da} show that there is not a direct connection between these errors and distance to camera or angle of view. We can see on Figures \ref{fig:scatter_ds} and \ref{fig:scatter_as} that the three missclassifications have a score below 0.85, highlighting the nature of this parameter as a confidence measure.

\begin{figure*}[h]
\centering
\begin{subfigure}{0.49\textwidth}
\centering
\includegraphics[width=\textwidth]{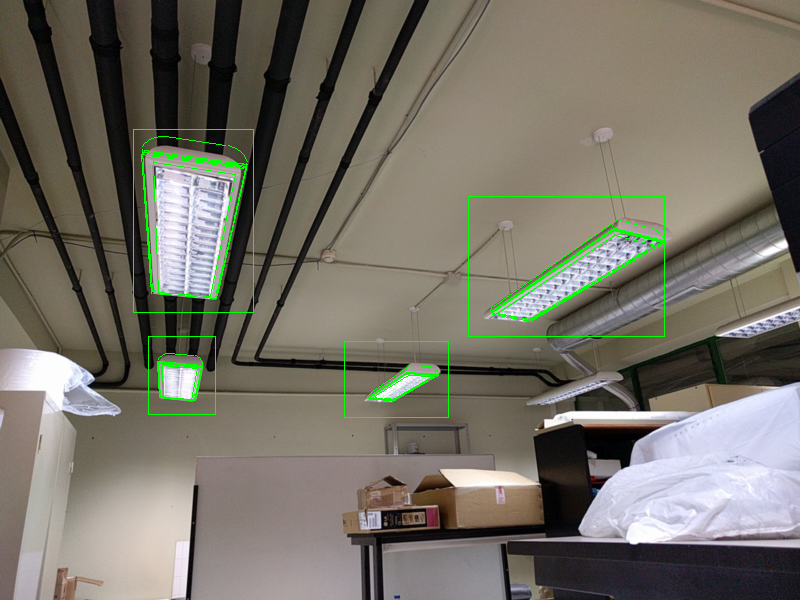}
\end{subfigure}
\begin{subfigure}{0.49\textwidth}
\centering
\includegraphics[width=\textwidth]{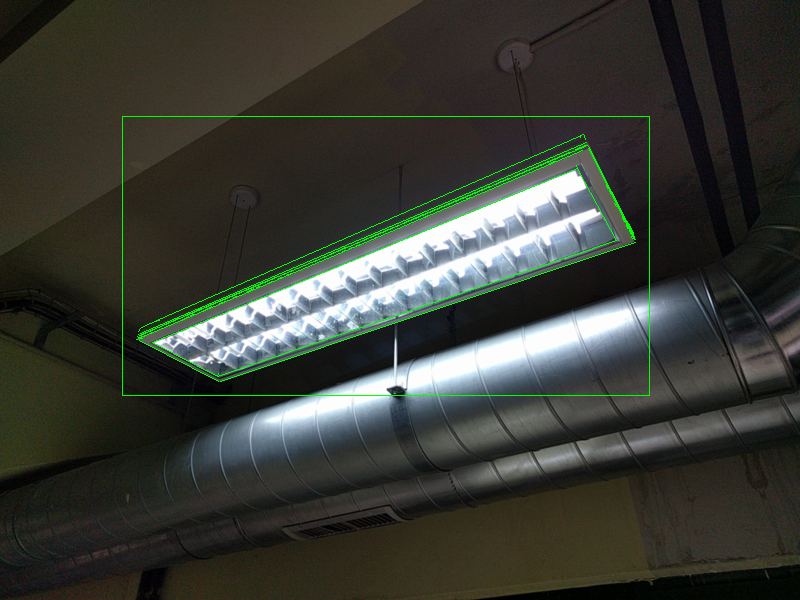}
\end{subfigure}

\begin{subfigure}{0.49\textwidth}
\centering
\includegraphics[width=\textwidth]{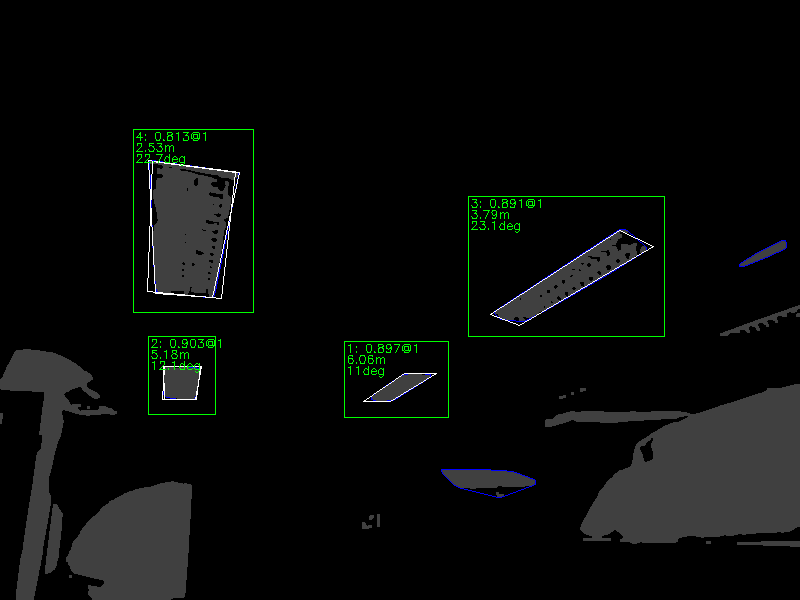}
\caption{Figure A}
\label{subfig:det_a1}
\end{subfigure}
\begin{subfigure}{0.49\textwidth}
\centering
\includegraphics[width=\textwidth]{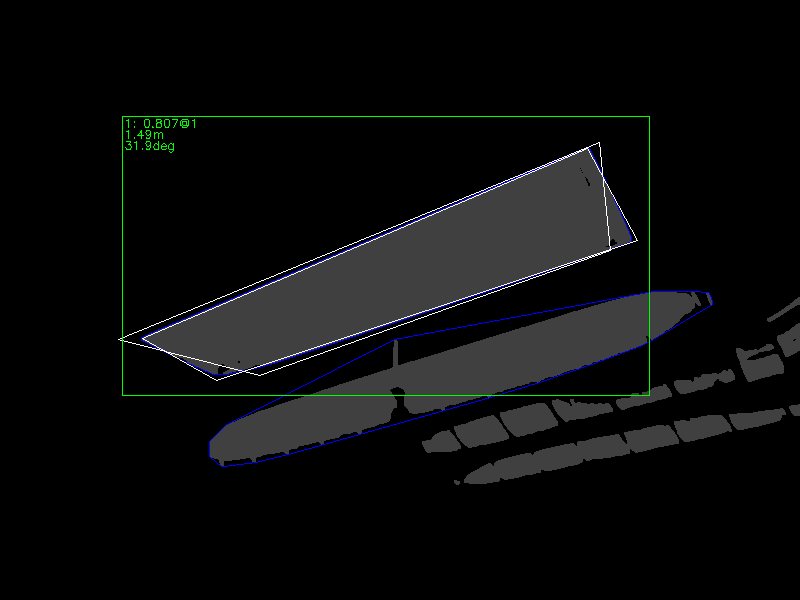}
\caption{Figure B}
\label{subfig:det_a2}
\end{subfigure}
\caption{Lamp detections and associated ROIs; up: color image, down: brightness mask.}
\label{fig:det_1}
\end{figure*}

\begin{figure*}[h]
\centering
\begin{subfigure}{0.49\textwidth}
\centering
\includegraphics[width=\textwidth]{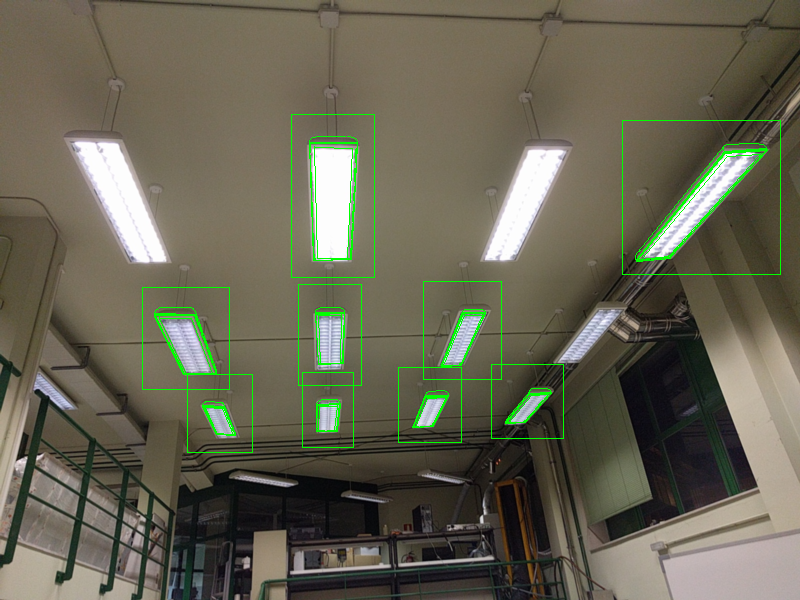}
\end{subfigure}
\begin{subfigure}{0.49\textwidth}
\centering
\includegraphics[width=\textwidth]{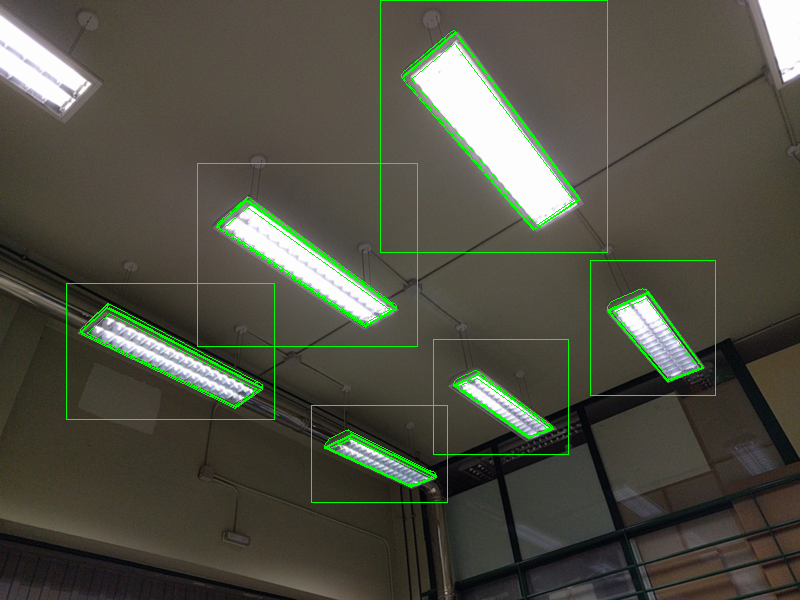}
\end{subfigure}

\begin{subfigure}{0.49\textwidth}
\centering
\includegraphics[width=\textwidth]{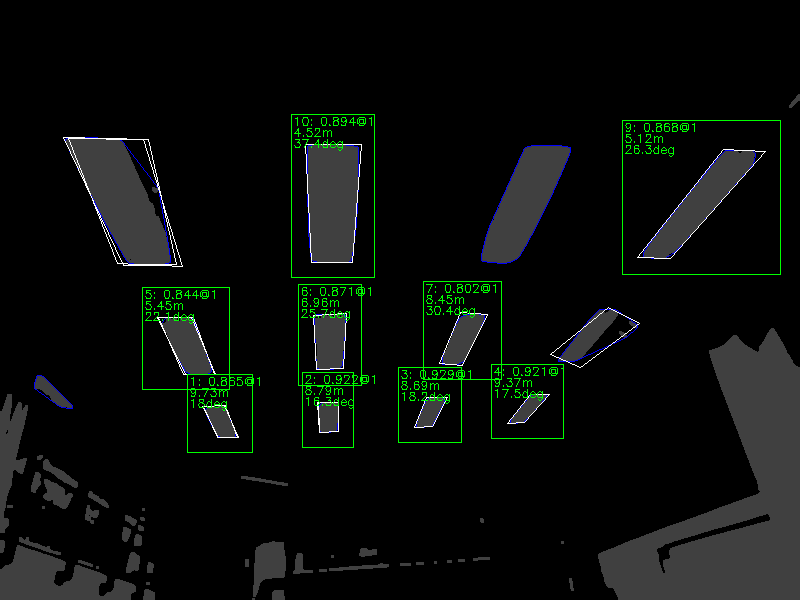}
\caption{Figure C}
\label{subfig:det_a3}
\end{subfigure}
\begin{subfigure}{0.49\textwidth}
\centering
\includegraphics[width=\textwidth]{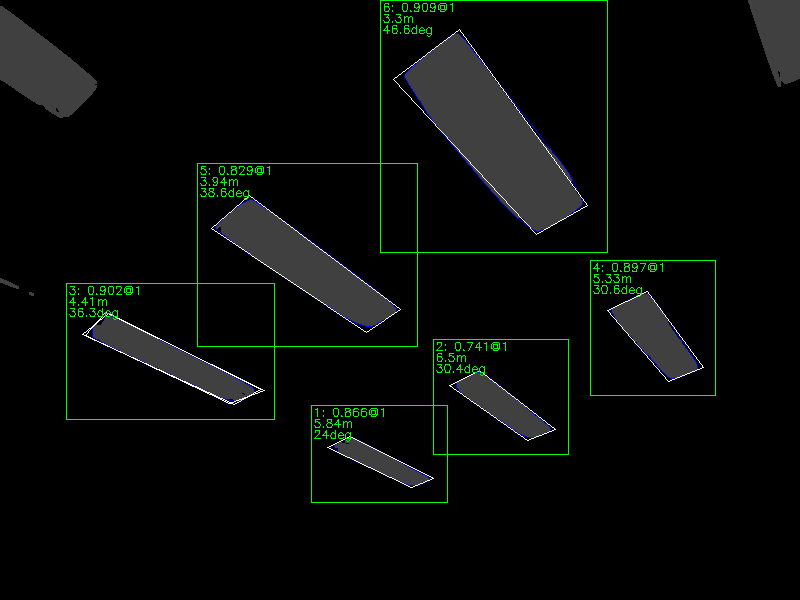}
\caption{Figure D}
\label{subfig:det_a4}
\end{subfigure}
\caption{Lamp detection and associated ROIs; up: color image, down: brightness mask.}
\label{fig:det_2}
\end{figure*}

\begin{figure*}[h]
\centering
\begin{subfigure}{0.49\textwidth}
\centering
\includegraphics[width=\textwidth]{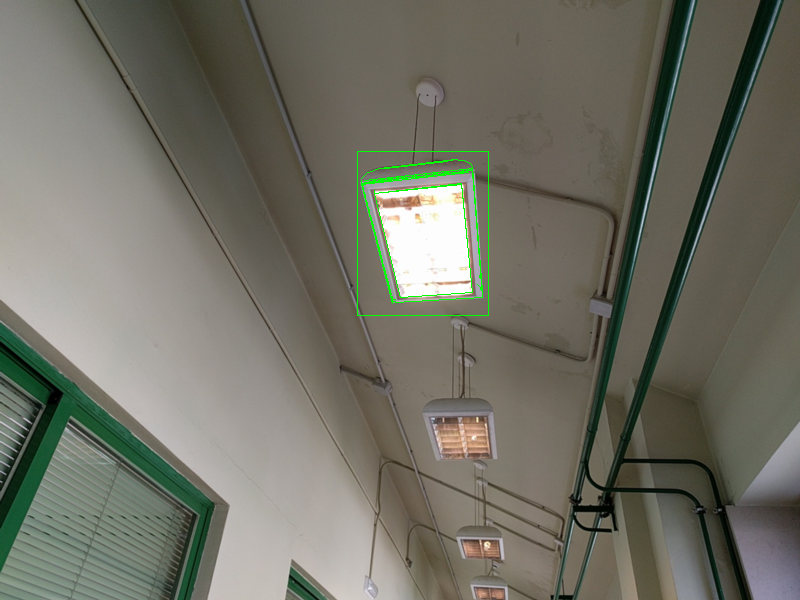}
\end{subfigure}
\begin{subfigure}{0.49\textwidth}
\centering
\includegraphics[width=\textwidth]{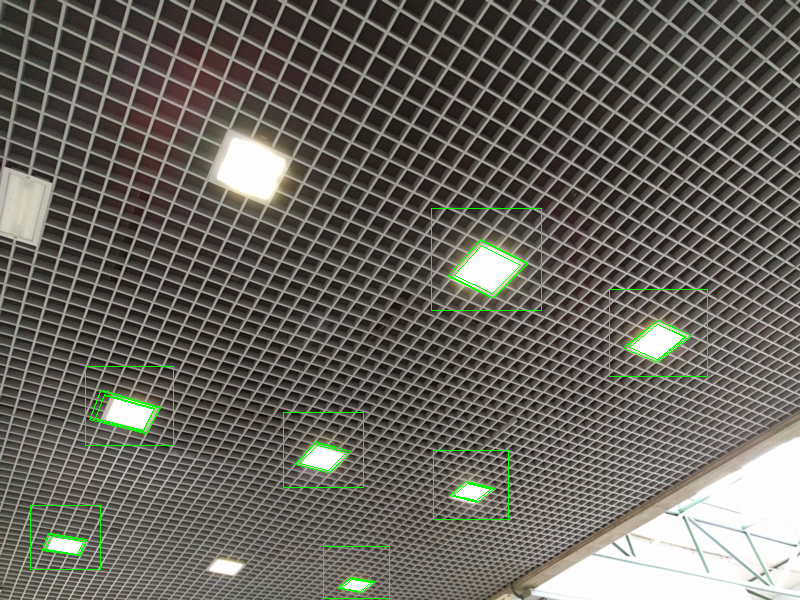}
\end{subfigure}

\begin{subfigure}{0.49\textwidth}
\centering
\includegraphics[width=\textwidth]{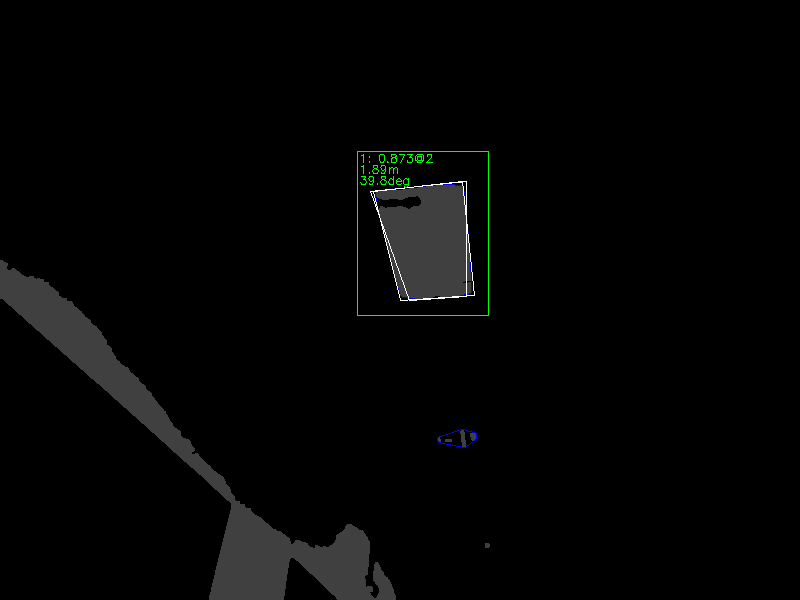}
\caption{Figure E}
\label{subfig:det_b}
\end{subfigure}
\begin{subfigure}{0.49\textwidth}
\centering
\includegraphics[width=\textwidth]{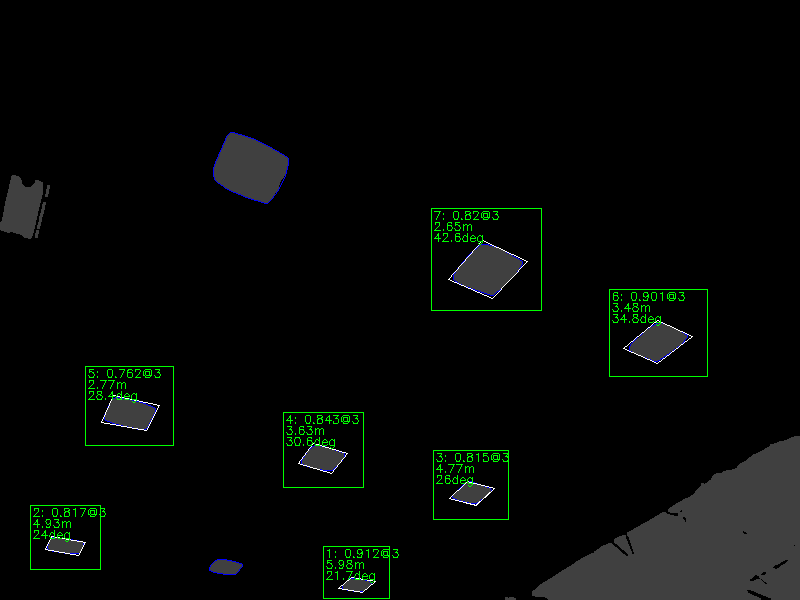}
\caption{Figure F}
\label{subfig:det_c}
\end{subfigure}
\caption{Lamp detections and associated ROIs; up: color image, down: brightness mask.}
\label{fig:det_3}
\end{figure*}

Figures \ref{fig:det_1} to \ref{fig:det_3} show some examples of lamp detection in different images. It is remarkable that the methodology detects the lamps even when the background is not clean and that the processing focuses only on small portions of the image.

The system can correclty classify lamps that are relatively far from the camera (as far as approx. 10~metres, as shown in Figure~\ref{subfig:det_a3}) and small angles (approx. 10~degrees from the ceiling plane, as shown in Figure~\ref{subfig:det_a1}). For larger distances and smaller angles, the light blobs become too small, and the edge information becomes too ambiguous for correct identification.

Furthermore, Figure~\ref{subfig:det_a2} shows that it correctly rejects bright reflections. Because this rejection is done in the first step of the algorithm, no region of interest is determined from the reflection, and consequently, no IDT3$_V$ is calculated, avoiding unnecessary computational costs. Finally, even when there are several close lamps in the ceiling, the regions of interest do not overlap so much, as presented in Figures \ref{subfig:det_a3} and \ref{subfig:det_a4}.

As shown in Figures \ref{subfig:det_b} and \ref{subfig:det_c}, the system correctly identifies the lamp model in the mayority of the test cases, even when model 2 is a shorter version of model 1 with very similar shape characteristics.

\subsubsection{Computational cost}
\label{subsubsec:ccost}

With the purpose of assessing the computational costs of our methodology, we have run all of the experiments on the same computer (featuring a Core i3-3240 CPU @ 3.40 GHz, 8 GB of RAM and Ubuntu 14.04 installed). We focus our study on the IDT3$_V$ computation, the slowest step, which will provide a better margin of improvement when applying our approach.

Our proposal simply avoids the calculation of the IDT3$_V$ for the entire image, which is only computed for a few regions of interest. The time needed in both scenarios (whole image computation, only specific regions of interest) is detailed in Tables \ref{tab:times} and \ref{tab:memory}, where it is easy to notice the great difference between the two alternatives. Our approach offers faster results and uses less memory, and even when the regions of interest slightly overlap, the total area is smaller than the entire image. Additionally, using different regions of interest supports a parallel implementation without modifying the tensor algorithm, which reduces the computational time for images with more than one region of interest even more so.

\begin{table*}[h]
\centering
\small
\caption{Processing time (in milliseconds) for the IDT3$_V$ computed over the whole image and computed only over the detected regions of interest. From left to right: 1) \textbf{Edge}: Edge extraction and tensor generation, 2) \textbf{Dist}: Distance transform, 3) \textbf{Rec}: Forward and backward recursions, 4) \textbf{Integ}: Integral images, 5) \textbf{Smth}: Smoothing for D$^2$CO, 6) \textbf{TOTAL}: Total tensor time, 7) \textbf{PAR}: Total tensor time with parallelization.}
\begin{tabular}{|c|l|r|r|r|r|r|r|r|}
\hline
\multicolumn{2}{|c|}{\textbf{Image}} & \textbf{Edge} & \textbf{Dist}
 & \textbf{Rec} & \textbf{Integ} & \textbf{Smth} & \textbf{TOTAL} & \textbf{PAR} \\
\hline
\multirow{2}{*}{\textbf{A}}
 & \textbf{Whole} & 61.66 & 80.66 & 108.05 & 77.10 & 56.82 & \textbf{384.38} & \textbf{-} \\
 & \textbf{ROIs}  & 11.40 & 20.85 & 10.14 & 13.36 & 11.70 & \textbf{68.01} & \textbf{34.06} \\
\hline
\multirow{2}{*}{\textbf{B}}
 & \textbf{Whole} & 66.72 & 80.78 & 107.82 & 77.20 & 56.95 & \textbf{389.55} & \textbf{-} \\
 & \textbf{ROIs}  & 28.99 & 27.77 & 24.51 & 22.63 & 19.09 & \textbf{125.68} & \textbf{-} \\
\hline
\multirow{2}{*}{\textbf{C}}
 & \textbf{Whole} & 60.17 & 80.44 & 107.57 & 76.94 & 56.91 & \textbf{382.12} & \textbf{-} \\
 & \textbf{ROIs}  & 15.58 & 44.23 & 19.06 & 27.42 & 25.43 & \textbf{132.98} & \textbf{59.03} \\
\hline
\multirow{2}{*}{\textbf{D}}
 & \textbf{Whole} & 58.47 & 80.78 & 107.34 & 76.75 & 56.84 & \textbf{380.26} & \textbf{-} \\
 & \textbf{ROIs}  & 19.84 & 41.76 & 27.35 & 28.83 & 22.51 & \textbf{142.43} & \textbf{82.78} \\
\hline
\multirow{2}{*}{\textbf{E}}
 & \textbf{Whole} & 88.29 & 80.28 & 107.73 & 76.84 & 56.84 & \textbf{410.07} & \textbf{-} \\
 & \textbf{ROIs}  & 3.13 & 5.75 & 3.46 & 5.05 & 3.27 & \textbf{20.88} & \textbf{-} \\
\hline
\multirow{2}{*}{\textbf{F}}
 & \textbf{Whole} & 84.11 & 80.36 & 107.99 & 77.09 & 56.91 & \textbf{406.57} & \textbf{-} \\
 & \textbf{ROIs}  & 8.79 & 19.08 & 7.97 & 11.66 & 12.22 & \textbf{60.36} & \textbf{25.07} \\
\hline
\end{tabular}
\label{tab:times}
\end{table*}

\begin{table*}[h]
\centering
\small
\caption{Memory usage (in MB) for a distance tensor computed over a whole image and computed only over the detected regions of interest.}
\begin{tabular}{|r|r|r|r|r|r|r|}
\hline
\textbf{Whole} & \textbf{A} & \textbf{B} & \textbf{C} & \textbf{D} & \textbf{E} & \textbf{F} \\
\hline
109.86 & 14.35 & 33.83 & 27.22 & 39.67 & 1.96 & 10.74 \\
\hline
\end{tabular}
\label{tab:memory}
\end{table*}

\begin{figure}[h]
\centering
\includegraphics[width=0.5\linewidth]{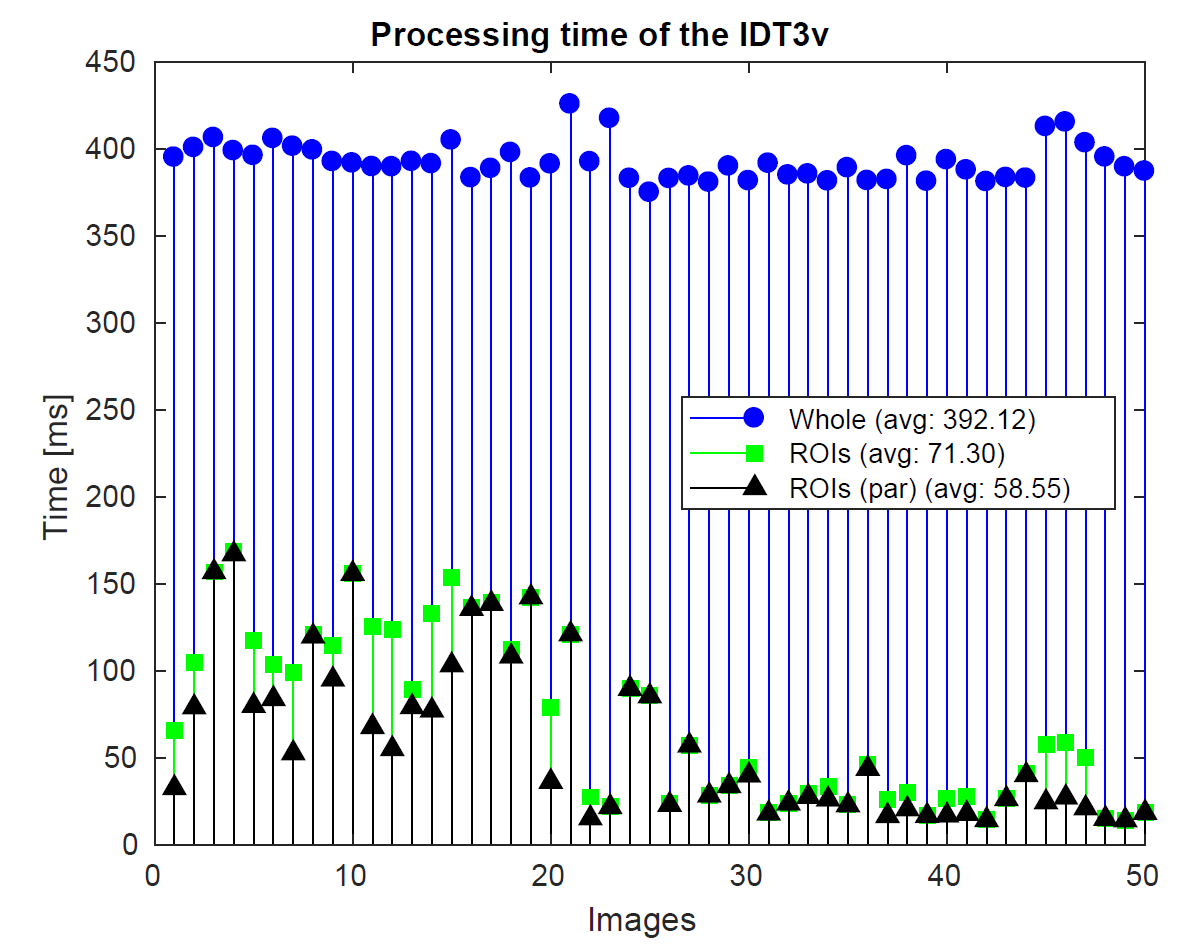}
\caption{Processing time (in millisenconds) of the IDT3$_V$ for each image in the dataset for the different configurations.}
\label{fig:times}
\end{figure}

We also include the time results for the dataset in Figure \ref{fig:times}. The average time for the IDT3$_V$ goes from $392.12$ ms to $71.30$ ms (18\%) with the use of ROIs and $58.55$ ms (15\%) if we further parallelize the processing in the tested images.

\subsection{Video and BIM}
\label{subsec:video}

\begin{figure}[h!]
\centering
\includegraphics[width=0.5\linewidth]{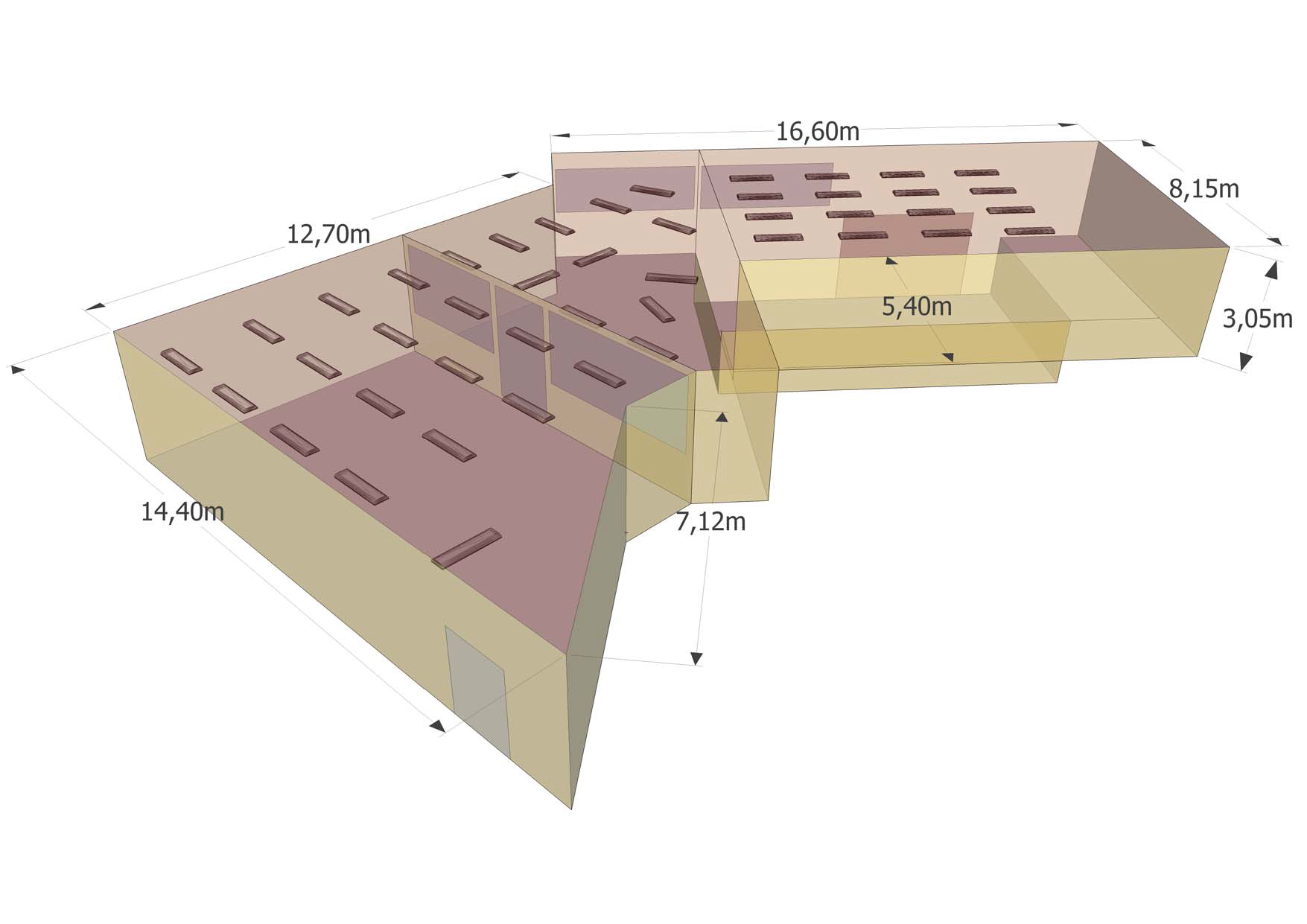}
\caption{3D model of the rooms used in the video tests.}
\label{fig:room}
\end{figure}

With the aim of assessing the performance or our methodology over video data, we have used a video sequence recorded in the Heat Engines and Machines laboratory at the University of Vigo. Figure~\ref{fig:room} shows the 3D model of the different rooms used in the experiments, which contain a total of 45 lamps of model 1.

To obtain localized images, we use an acquisition process analogous to the one presented by Lag\"{u}ela \emph{et al.}~\cite{Laguela}. The position and orientation of the camera are estimated using an inertial measurement unit and LiDAR sensors that output around 2 pose values per second, whereas the images are obtained using a $640\times480$ pixel webcam at 30 frames per second. Both inputs are combined via timestamps registered for each value, interpolating the poses for each video frame. The acquisition system is mounted on a backpack carried by the operator, so there are some errors in the poses, specially for frames that are not close to a pose detection.

\begin{figure*}[h!]
\centering
\begin{subfigure}{0.235\textwidth}
\centering
\includegraphics[width=\textwidth]{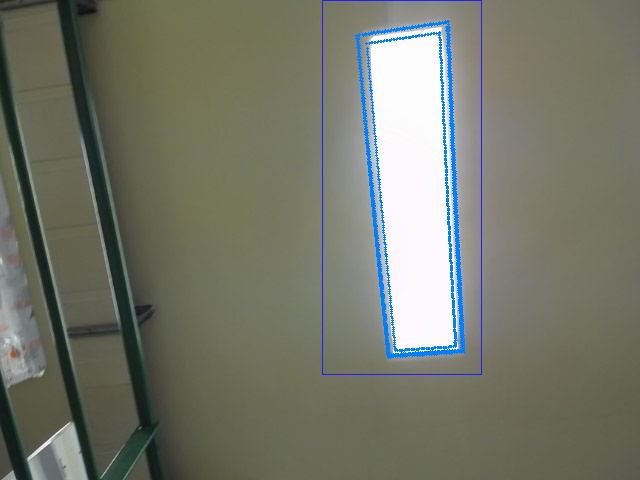}
\caption{}
\label{subfig:vid_0a}
\end{subfigure}
\begin{subfigure}{0.235\textwidth}
\centering
\includegraphics[width=\textwidth]{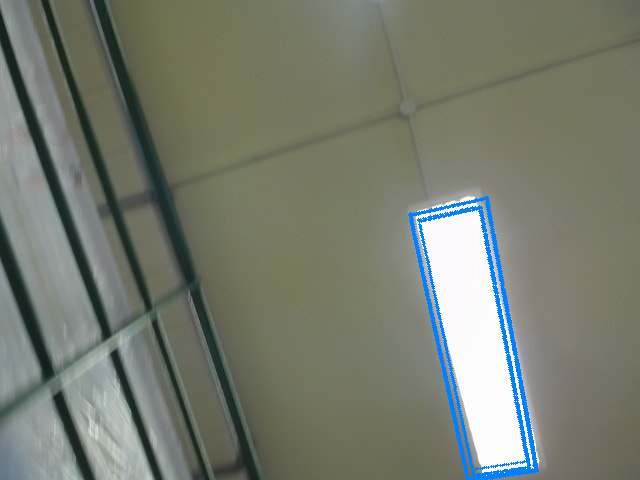}
\caption{}
\label{subfig:vid_0b}
\end{subfigure}
\begin{subfigure}{0.235\textwidth}
\centering
\includegraphics[width=\textwidth]{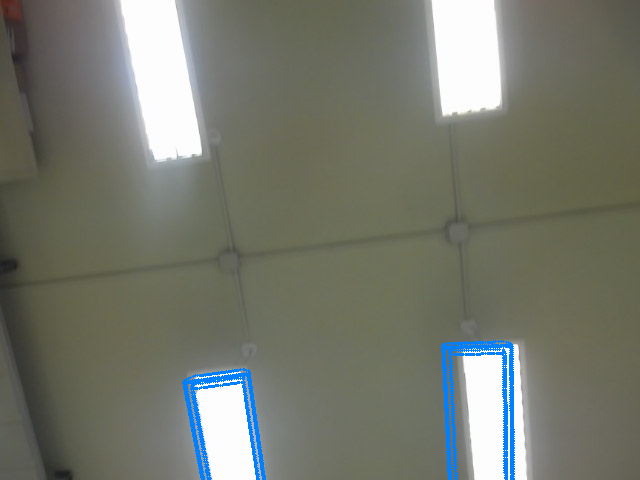}
\caption{}
\label{subfig:vid_2a}
\end{subfigure}
\begin{subfigure}{0.235\textwidth}
\centering
\includegraphics[width=\textwidth]{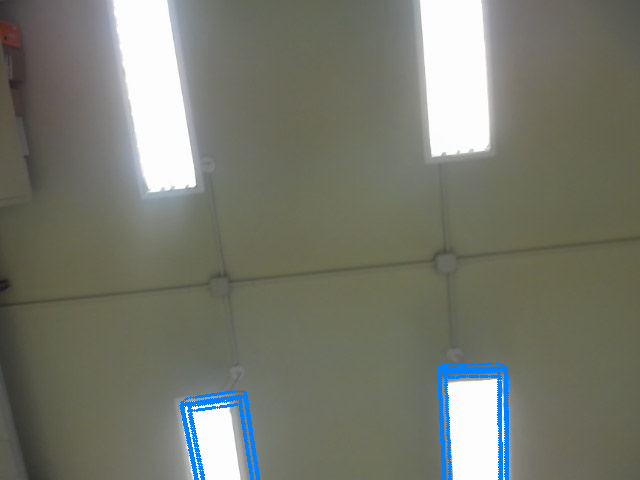}
\caption{}
\label{subfig:vid_2b}
\end{subfigure}

\begin{subfigure}{0.235\textwidth}
\centering
\includegraphics[width=\textwidth]{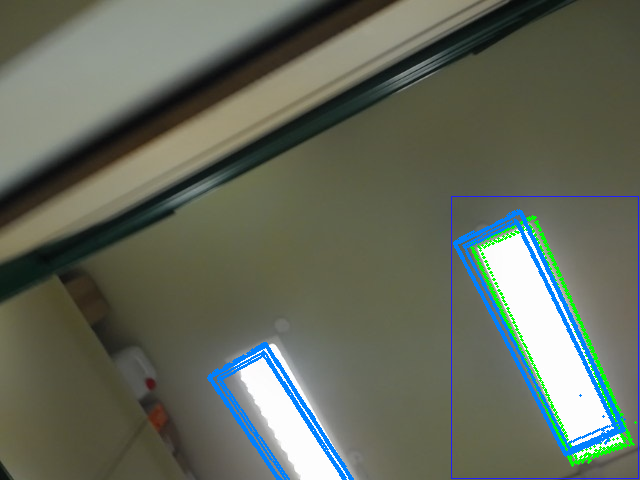}
\caption{}
\label{subfig:vid_3a}
\end{subfigure}
\begin{subfigure}{0.235\textwidth}
\centering
\includegraphics[width=\textwidth]{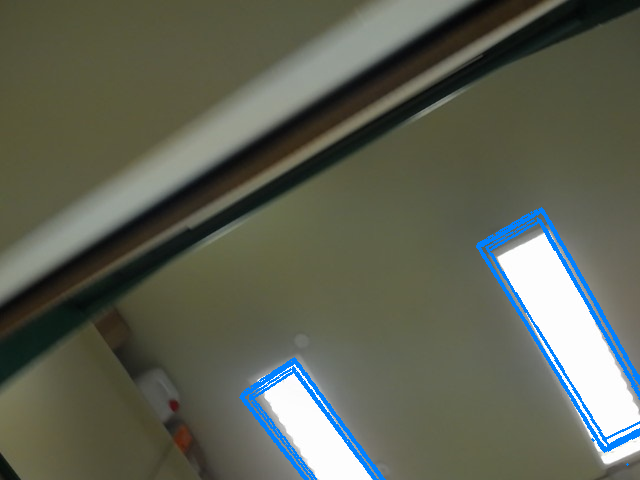}
\caption{}
\label{subfig:vid_3b}
\end{subfigure}
\begin{subfigure}{0.235\textwidth}
\centering
\includegraphics[width=\textwidth]{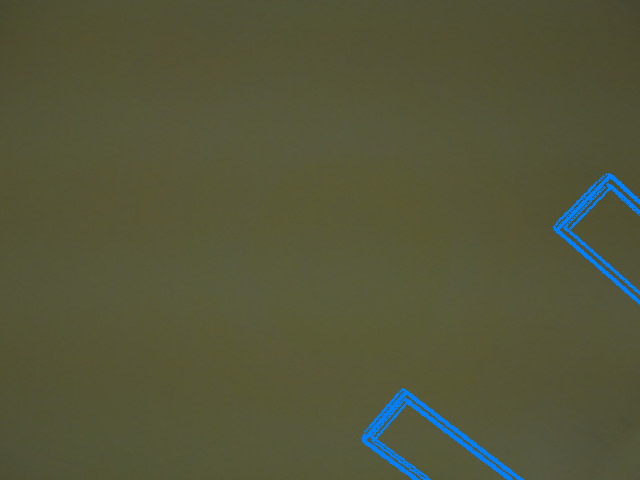}
\caption{}
\label{subfig:vid_3c}
\end{subfigure}
\begin{subfigure}{0.235\textwidth}
\centering
\includegraphics[width=\textwidth]{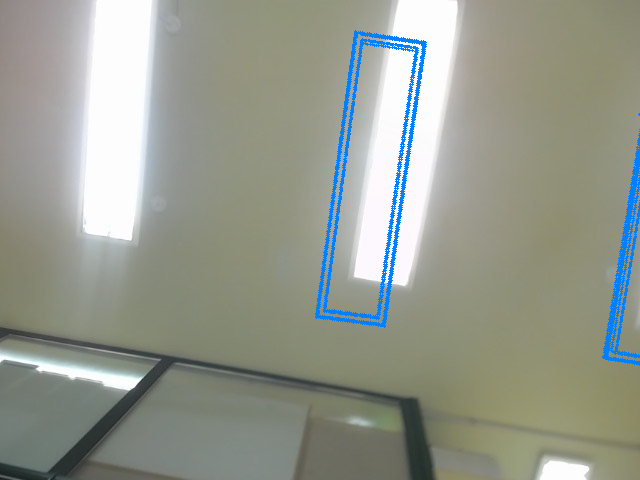}
\caption{}
\label{subfig:vid_3d}
\end{subfigure}
\caption{Detections extracted from different frames of the test sequence.}
\label{fig:vid}
\end{figure*}

Figure~\ref{fig:vid} shows some video frames from the recorded data: Image \ref{subfig:vid_0a} shows the first detection, where there is a little displacement error on a subsequent frame (Image \ref{subfig:vid_0b}), but it is small enough for the new detection to not be incorrectly registered as a different object. Images \ref{subfig:vid_2a} and \ref{subfig:vid_2b} show the same phenomenon for other detected lamps as well. Even when the original object completely leaves the camera coverage and there is a severe location and orientation change (Images \ref{subfig:vid_3a} to \ref{subfig:vid_3c}), the registered poses still lie close to the ones seen in later frames (Image \ref{subfig:vid_3d}).

In this case we have only registered two thermal zones. If we want to scan entire floors or buildings the errors from the odometry system may have an impact on the scalability of the method depending on the characteristics and the extension of the building. There are methods, like visual odometry, that could be applied in such situations to improve the accuracy of the recorded camera poses.

\begin{figure}[h!]
\centering
\includegraphics[width=0.4\textwidth]{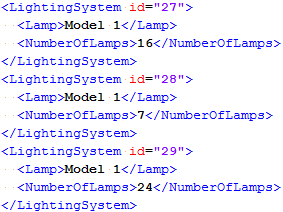}
\caption{Green Building XML lighting information generated by the detection system.}
\label{fig:gbxml_out}
\end{figure}

The registered detections can be used to automatically include lighting information in the BIM of the building. In fact, each one includes the associated lamp model, so we can fill in \texttt{LightingSystem} elements that are linked to each \texttt{Space} in the building. Figure \ref{fig:gbxml_out} shows a piece of gbXML with the most basic information generated from the detection system.

\section{Conclusions}
\label{sec:conclusions}

We have presented a robust system for the detection, identification and localization of lamps in building interiors that combines a novel method for candidate search with state-of-the-art methods for candidate selection and refinement. The three stage system was applied using three different lamp models on a set of 51 images containing 96 detected lamps extracted from three sequences, ensuring different camera conditions with a pose distance of at least 50 cm in position or 20$^\circ$ in orientation.

Using brightness and shape information from the lamp models, our method performs a very efficient search, which greatly reduces the computational needs required for object localization. In particular, the computational time was reduced to 18\% (15\% with further parallelization) of the original time on average in the most expensive step in our tests.

Moreover, the method can identify lamps up to 10 metres, classifying 96.9\% of the detections correctly in this range in our experiments, which makes it suitable for lamp detection in both residential and industrial buildings. Our video experiment assessed the validity of the method in a continuous operation, avoiding duplicate detections for the same object.

As shown, the resulting information can be easily added to the energy model of the building, allowing automatic determination of the lighting information and accelerating the definition of the BIM.

We plan to test our system with more lamp models in the future. Additionally, we are exploring the use of thermal imaging to extend the search to other building elements, e.g., thermal installations and thermal bridges.

\section{Acknowledgements}

Authors want to give thanks to the Xunta de Galicia (Grant ED481A) and the European Union's Horizon 2020 research and innovation programme under grant agreement No. 720661.


\bibliographystyle{ieeetr}
\bibliography{Biblio}

\end{document}